\documentclass[a4paper]{article}

\usepackage{authblk,rotating}
\usepackage{amsmath}
\usepackage{subcaption}
\usepackage{amssymb}
\usepackage{tikz}
\usepackage{overpic}    
\usepackage[vlined,ruled]{algorithm2e}
\SetKwInOut{Input}{input}
\SetKwInOut{Output}{output}
\SetKwComment{Comment}{}{}

\SetCommentSty{mycmtsty}
\usepackage{tikz}
\usetikzlibrary{spy}
\usepackage{amsthm}

\newtheorem{remark}{Remark}
\newtheorem{theorem}{Theorem}[section]
\newtheorem{proposition}{Proposition}[section]

\def\R{{\mathbb R}}

\newcommand{\argmin}{\mathop{\mathrm{Argmin}}}
\newcommand{\argmax}{\mathop{\mathrm{Argmax}}}

\newcommand{\im}{\mathrm{Im}}
\newcommand{\id}{\mathrm{Id}}

\newcommand{\psnr}{\mathsf{PSNR}}
\newcommand{\fix}{\mathsf{Fix}}

\newcommand{\prox}[1]{\mathsf{Prox}_{#1}}

\usepackage{fancyvrb}
\VerbatimFootnotes 

\SetKwInput{Param}{Param.}
\title{From the Gradient-Step Denoiser to the Proximal Denoiser and their associated convergent Plug-and-Play algorithms}

 \author[1,2,3]{Vincent Herfeld}
 \author[1]{Baudouin Denis de Senneville}
 \author[4]{Arthur Leclaire} 
 \author[1]{Nicolas Papadakis}

 \affil[1]{Univ. Bordeaux, CNRS, INRIA, Bordeaux INP, IMB, UMR 5251, F-33400 Talence, France}
 \affil[2]{Université Grenoble Alpes, Inria, CNRS, Grenoble INP, LJK, France}
 \affil[3]{Enhance Lab,  Paris, France}
 \affil[4]{LTCI, Télécom Paris, IP Paris}

\date{}
\begin{document}

\maketitle

\begin{abstract}
In this paper we analyze the Gradient-Step Denoiser and its usage in Plug-and-Play algorithms. The Plug-and-Play paradigm of optimization algorithms uses off the shelf denoisers to replace a proximity operator or a gradient descent operator of an image prior. Usually this image prior is implicit and cannot be expressed, but the Gradient-Step Denoiser is trained to be exactly the gradient descent operator or the proximity operator of an explicit functional while preserving state-of-the-art denoising capabilities.
\end{abstract}

\section{Introduction\label{intro}}

This paper is dedicated to a practical study of Plug-and-Play algorithms for imaging inverse problems, based on the gradient-step denoiser~\cite{hurault2021gradient} or the proximal denoiser~\cite{ProxPnP}. 
Solving inverse problems consists in recovering a clean signal $x_*$ from a noisy observation $y = A x_0 + w$ where $A$ is a linear operator, $x_0$ is the clean original image we wish to recover and $w$ is a white Gaussian noise of variance $\sigma^2$.
Here, we will consider imaging inverse problems where $A$ is for example a blurring operator. 
These problems are ill-posed in general, meaning that there exist several solutions, possibly infinitely many, and where only a few correspond to interesting images. 
To cope with that, we can regularize the problem, and then  minimize objective functions of the form
\begin{equation}
    F = \frac1\lambda f + g ,
    \label{ref:energy}
\end{equation}
where $f$ is the data fidelity term to the observation $y$, $g$ a regularizing prior, and $\lambda \in \R_+$ a parameter that sets the balance between fidelity and prior knowledge. The solution set corresponds to the minimizers of this objective function~$F$. 

This kind of minimization problem can be tackled with proximal splitting algorithms, such as Proximal Gradient Descent (PGD), Half-Quadratic Splitting (HQS) or the Douglas-Rachford algorithm (DRS). 
In \cite{hurault2021gradient,ProxPnP} such algorithms leverage on the gradient-step operator, or the proximity operator which are defined for a given function $g : \R^n \to \R$ respectively as $\id - \nabla g$ (provided $g$ is differentiable) and
\begin{equation}
\prox{g}(x) = \argmin_{y \in \R^n} g(y) + \frac{1}{2}||x - y||^2 \subset \R^n .
\end{equation}
$\prox{g}$ is a point-to-set operator, which is single-valued as soon as $g$ is convex proper and lower semi-continuous. 
We say that a function is \textit{proximable} if its proximity operator can be computed in closed form. 
These splitting algorithms correspond to fixed point iteration schemes $x_{k+1} = T(x_k)$ applied with different operators:
\begin{align}
    \textbf{(PGD) } T^{\tau }_{PGD} &= \prox{\tau g} \circ \left(\id - \frac{\tau}\lambda   \nabla f\right)\\
    \textbf{(HQS) } T^{\tau}_{HQS} &= \prox{\tau  g }\circ  \prox{\frac{\tau}\lambda  f }\\
    \textbf{(DRS) } T^{\tau}_{DRS} &= \beta (\mathsf{RProx}_{\tau  g} \circ \mathsf{RProx}_{\frac{\tau}\lambda f}) +(1 - \beta) \id,
\end{align}
where $\tau$ is a step size and $\mathsf{RProx}_f$ is the reflected proximity operator of $f$, defined by
\begin{equation}
    \mathsf{RProx}_f := 2 \prox{f} - \id.
\end{equation}

Many convergence results for proximal splitting algorithms have been obtained, and they still constitute an important research field today. 
These results rely on hypotheses on $f$ and $g$ such as convexity, differentiability, proximability, smoothness (e.g. having Lipschitz gradient). 
Plug-and-Play (PnP) algorithms, introduced in~\cite{PnP} and~\cite{RED} form a new application of these proximal algorithms, where a pre-trained denoiser is used as the proximal operator or the gradient of the regularization $g$.
\begin{remark}
  One intuition behind the regularization of inverse problems is given by a Bayesian model. 
    Using a noise model $p(y|x)$, a prior $p_X(x)$ on $x$, and Bayes formula, the search for the maximum-a-posteriori consists in solving
    \begin{align*}
        \argmax_{x \in \mathcal{X}} p(x | y) 
        =  \argmax_{x \in \mathcal{X}} \frac{p(y | x)p_X(x)}{p(y)} 
        = \argmin_{x \in \mathcal{X}} \underbrace{-\log p(x | y)}_{\text{data-fidelity }f(x)} \; \underbrace{-\log p_X(x)}_{\text{log prior }g(x)}.
    \end{align*}
    For example for an additive white Gaussian noise $w \in \mathcal{N}(0, \sigma^2 \id)$, $f(x) = \frac{1}{2 \sigma^2}||Ax - y||^2$. 
\end{remark}

As of the prior regularizer, either we consider priors related to explicit regularizations such as Tikhonov~\cite{Tikhonov:1963}, Total Variation~\cite{TV}, Wavelet regularization~\cite{wavelet} etc.
Or, as it is done in Plug-and-Play algorithms, we consider an implicit regularization associated to a Gaussian Denoiser. 
There exist two main theoretically ideal denoisers, the MMSE (Minimum Mean Squared Error) denoiser and the MAP (maximum-a-posteriori) denoiser.
For a white Gaussian noise of variance $\sigma^2$, the MAP denoiser is the proximal operator of the regularization $g$~\cite{MAP}:
\begin{equation} \label{D_MAP} 
    D^{MAP}_\sigma = \prox{-\sigma^2 \log p_X} = \prox{\sigma^2g} .
\end{equation}
Also, it can be shown that the MMSE denoiser is related to the smoothed log-prior through Tweedie identity~\cite{Tweedie}:
\begin{equation} \label{D_MMSE}
    D^{MMSE}_\sigma = \text{Id } + \sigma^2 \nabla \log p_\sigma , 
\end{equation}
where $p_\sigma = p_X * \mathcal{N}(.; 0, \sigma^2 \id)$ is the convolution of $p_X$ with a Gaussian density of variance $\sigma^2 \id$.

One objection of the Plug-and-Play methodology is that the used off-the-shelf (pre-trained) denoisers may not correspond to true MMSE or MAP denoisers.
This impacts convergence results as well as the quality of the obtained solutions. To make Plug-and-Play more robust to these issues, several works propose to add additional assumptions on the denoiser that are milder than those from the original works of~\cite{PnP} and~\cite{RED}, but stay difficult to verify in practice~\cite{RED-Clar}. These works suggest to control the denoiser training~\cite{PnP-PTD} as well as to consider restrictive theoretical assumptions such as firm nonexpansivity~\cite{Scalable-PnP}, averaged denoiser~\cite{PnP-eig} and others.

The main contribution of~\cite{hurault2021gradient,ProxPnP} is to propose convergence guarantees of Plug-and-Play algorithms, by using a simple modification of the denoising network. The considered Gaussian Denoiser, called Gradient-Step Denoiser (GS), is defined and trained such as it verifies  one of the above identities, \eqref{D_MAP} or \eqref{D_MMSE}. This formulation also makes the prior $g$ explicit, meaning it can be computed. In addition to guaranteed convergence, we are thus able to evaluate $F$ and monitor convergence along the iterations.

\paragraph{Outline}
In Section~\ref{sec:GS-PnP} we will present the Gradient-Step Plug-and-Play (GS-PnP) framework.  
Section~\ref{sec:prox-PnP} will be dedicated to the extension of the Gradient Step framework which is called Proximal Plug-and-Play (Prox-PnP). 
Section~\ref{sec:Training} aims to present practical details on the denoiser training. 
In Section~\ref{sec:Exp}, we provide several simulations obtained with these gradient-step and proximal Plug-and-Play algorithms, as well as a thorough analysis of their main parameters.
\section{Gradient-Step Plug-and-Play}\label{sec:GS-PnP}
In Section \ref{GSD}, we first explain the Gradient-Step denoiser and how it is trained. 
Then in Section~\ref{sec:GS-PnP alg} we present the GS-PnP algorithm, and in Section~\ref{sec:gspnp_cv} we recall the main theoretical results that ensure its convergence. 

\subsection{The Gradient-Step Denoiser}\label{GSD}
In~\cite{hurault2021gradient}, a denoising operator $D_\sigma$ is introduced to act as a gradient descent step over an explicit differentiable function $g_\sigma : \mathbb{R}^n \rightarrow \mathbb{R}$: 
\begin{equation}\label{def:denoiser}
    D_\sigma = \text{Id } - \nabla g_\sigma.
\end{equation}
The parameter $\sigma > 0$ refers to the variance of the Gaussian noise used in the training of the denoiser~$D_\sigma$.
There may be several choices of $g_\sigma$ that lead to good denoising results.
Then it is proposed in ~\cite{hurault2021gradient} not to directly parameterize $g_\sigma$ by a scalar neural network (which resulted in poor denoising capabilities) but to consider 
\begin{equation}\label{eq:gsigma}
g_\sigma(x) = \frac{1}{2} ||x - N_\sigma (x)||^2, 
\end{equation}
where $N_\sigma$ is parameterized by a denoising differentiable neural network. This gives
\begin{equation}
    D_\sigma(x) 
    = N_\sigma(x) + J_{N_\sigma}(x)^T (x - N_\sigma(x)) \label{eq:D_sig}
\end{equation}
where $J_{N_\sigma}(x)$ is the Jacobian of $N_\sigma$ at $x$ computed via auto-differentiation. 
Compared to the RED framework~\cite{RED}, we see here that the denoiser $D_\sigma$ is a gradient field, (that is, a conservative vector field), which is important for convergence analysis.

The application of the GS denoiser differs from classical methods that just realizes a forward pass in a neural network. As we can see in equation \eqref{eq:D_sig}, here computing $D_\sigma(x)$ requires  a forward pass to compute $N_\sigma(x)$ and a backward pass for $\id - \nabla g_{\sigma}(x)$.
\paragraph{Training loss }\label{sec:train1}
For a noise variance $\sigma^2$, the denoiser $D_\sigma$ is trained to approximate the MMSE denoiser by minimizing the following loss, over the data distribution $p_{data}$:
\begin{align}
    \mathcal{L}(D_\sigma) &= \mathbb{E}_{x \sim p_{data}, \xi_\sigma \sim \mathcal{N}(0, \sigma^2 I)}\left[ ||D_\sigma(x + \xi_\sigma) -  x||^2\right]\nonumber\\
    &= \mathbb{E}_{x \sim p_{data}, \xi_\sigma \sim \mathcal{N}(0, \sigma^2 I)}\left[ ||\nabla g_\sigma(x + \xi_\sigma) - \xi_\sigma||^2\right]\hspace{-1pt}.\label{loss-gsd}
\end{align} 
\subsection{The GS-PnP algorithm}\label{sec:GS-PnP alg}
We now present the Gradient-Step Plug-and-Play algorithm used to solve the optimization problem
\begin{equation}\label{ref:gs-energy}
    \argmin_{x \in \mathbb{R}^n} F(x) = \frac{1}{\lambda } f(x) + g_\sigma(x)
\end{equation}
where $g_\sigma$ is the regularization defined by~\eqref{eq:gsigma}.

Plug-and-Play consists in replacing the proximity operator or the gradient step operator of a function by a denoising operator. For instance the PnP-PGD operator 
$D_\sigma \circ (\id - \tau \nabla f )$
replaces the original PGD (Proximal Gradient Descent) operator 
$\prox{\tau g }\circ  (\id - \tau \nabla f )$, where $\tau \geq 0$ corresponds to the algorithm step-size.  In~\cite{hurault2021gradient}, as defined in equation~\eqref{def:denoiser}, the denoiser is taken as a gradient descent operator. The roles of $f$ and $g$ are inverted and one obtains the GS-PnP operator: 
\begin{align}
\begin{split}
    T^{\tau, \lambda}_{GS-PnP} = \prox{\frac{\tau}{\lambda} f} \circ (\text{Id } - \tau \nabla g_\sigma) 
    &= \prox{\frac{\tau}{\lambda} f} \circ (\tau (\text{Id } -\nabla g_\sigma) + (1 - \tau )\id)\\
    &= \prox{\frac{\tau}{\lambda} f} \circ (\tau  D_\sigma + (1 - \tau )\id).
\end{split}
\end{align}
This operator is used in a fixed point scheme such that the GS-PnP algorithm corresponds to:
\begin{align}\label{GSPnP-Scheme}
\begin{split}
    x&_0 \in \mathbb{R}^n,\\
    x&_{k+1} = T^{\tau, \lambda}_{GS-PnP}(x_k), \; \forall k \geq 0.
\end{split}
\end{align}

\subsection{Convergence analysis}
\label{sec:gspnp_cv}
We now present the main theorems established in the  paper~\cite{hurault2021gradient}, that contains all convergence proofs. 

\begin{theorem}[Theorem 1 in~\cite{hurault2021gradient}] \label{Thm : 1}
    Let $f :\mathbb{R}^n \rightarrow \mathbb{R}\cup \{+\infty\}$ and $g_\sigma : \mathbb{R}^n \rightarrow \mathbb{R}$ be proper lower
semicontinous functions with $f$ convex and $g_\sigma$ differentiable with $L$-Lipschitz gradient. Let $\lambda > 0$,
$F = \frac{1}{\lambda}f + g_\sigma$ and assume that $F$ is bounded from below. Then, for $\tau < \frac{\lambda}{L}$, the iterates $x_k$ given by the iterative scheme (\ref{GSPnP-Scheme}) verify
\begin{enumerate}
    \item $(F(x_k))$is non-increasing and converges.
    \item The residual $|| x_{k+1} - x_k ||$ converges to $0$.
    \item All cluster points of the sequence $(x_k)$ are stationary points of (\ref{ref:gs-energy}).
\end{enumerate}
\end{theorem}

\begin{remark}\label{rem:lipschitz}
  As can be seen in~\cite{hurault2023thesis}, by choosing activation functions of $N_\sigma$ that are Lipschitz and with Lipschitz gradient, it becomes reasonable to assume that  $\nabla g_\sigma$ is $L$-lipschitz for some constant $L$.
\end{remark}

For the second theorem we need an additional technical property on functions for convergence proofs in non convex optimization setting. We will thus consider the Kurdyka-Lojasiewicz (KŁ) property and 
we refer to~\cite{KL} for its precise definition, which is not of crucial interest here.
The class of KŁ functions is very large and the regularization functions that  we will used later are designed to satisfy this KŁ property (we refer to~\cite{hurault2023thesis} for the details of the proof).

\begin{theorem}[Theorem 2 in~\cite{hurault2021gradient}] \label{Thm : 2}
    Let $f :\mathbb{R}^n \rightarrow \mathbb{R}\cup \{+\infty\}$ and
$g_\sigma : \mathbb{R}^n \rightarrow \mathbb{R}$ be proper lower semicontinous functions with $f$ convex and $g_\sigma$ differentiable with $L$-Lipschitz gradient. Let $\lambda > 0$,
$F = \frac{1}{\lambda} f + g_\sigma$ and assume that $F$ is bounded from below. Assume
that $F$ verifies the KŁ property. Suppose that $\tau < \frac{\lambda}{ L}$ . If the sequence $(x_k)$ given by the iterative
scheme \eqref{GSPnP-Scheme} is bounded, then it converges, with finite length, to a critical point $x^*$ of $F$ .
\end{theorem}
An important assumption for Theorem~\ref{Thm : 2} to be verified is the boundedness of the iterates. Since we consider lower semicontinuous functions a sufficient condition for $(x_k)$ to be bounded is for $F$ to be coercive: $\lim_{||x|| \rightarrow +\infty}F(x) = +\infty$. Indeed thanks to the sufficient decrease property explained below in Section \ref{backtracking} we always have $F(x_k) \leq F(x_0)$ and so $||x_k||$ is never going to $\infty$. 

In practice, it is proposed in~\cite{hurault2021gradient} to slightly modify the regularization and enforce the coercivity of $F$ by adding a term that will recall the iterates to a convex set (chosen to be $C = [-1, 2]^n$). From
\begin{align}
    \tilde{g}_\sigma(x) = g_\sigma(x) + \frac{1}{2}||x - \Pi_C(x)||^2_2  =  \frac{1}{2}||x - N_\sigma(x)||^2_2 + \frac{1}{2}||x - \Pi_C(x)||^2_2 
\end{align}
which is still differentiable, we get the modified gradient-step
\begin{align}
    \tilde{D}_\sigma(x) = (x - \nabla g_\sigma(x)) - (x - \Pi_C(x)).
\end{align}
It is mentioned in~\cite{hurault2021gradient} that, even though this projection has been implemented, it does not activate in practice. In our implementation we use this remark to avoid useless  projection steps. 
\subsection{Backtracking}\label{backtracking}
As seen in Section~\ref{sec:gspnp_cv}, the convergence guarantee for the GS-PnP algorithm relies on the assumption that $g_\sigma$ (used in \eqref{ref:gs-energy}) is $L$-smooth, meaning differentiable with $L$-Lipschitz gradient, and that the time-step $\tau$ verifies the inequality $\tau < \frac{\lambda}{L}$. Estimating $L$ over a subset of the iterates $(x_k)$ does not lead to optimal convergence speed.
As it is explained in~\cite{Beck}, using a backtracking procedure allows to dynamically change the step-size so that the sufficient decrease property is verified at each iteration.
The following condition corresponds to this idea and is part of the GS-PnP algorithm:
\begin{align}
    \textbf{While } F(x_k) - F(T^{\tau, \lambda}_{GS-PnP}(x_k)) < \frac{\gamma}{\tau} ||x_k - T^{\tau, \lambda}_{GS-PnP}(x_k)||^2_2, \; \tau \leftarrow \eta \tau.
\end{align}
 It is also proven that this while loop terminates in a finite number of steps.
 
 The GS-PnP pseudo-code is summarized in Algorithm~\ref{alg:GS-PnP}.
\begin{algorithm}[!ht]
\caption{Gradient-Step Plug-and-Play (GS-PnP)}
\DontPrintSemicolon
\Param{init. $z_0,\lambda > 0, \sigma > 0, \epsilon > 0, \tau_0 > 0, K \in \mathbb{N}^*, \eta \in (0, 1), \gamma \in (0, 1/2)$}
\Input{degraded image $y$}
\Output{restored image $\hat{x}$}
$k = 0, x_0 = \prox{\frac{\tau}{\lambda} f}(z_0), \tau = \tau_0 / \eta, \Delta > \epsilon$\\
\While{$k < K$ and $\Delta > \epsilon$}{
  $z_k = \tau D_\sigma(x_k) + (1 - \tau)x_k$ \Comment*{Apply Gradient Step on previous iteration}
  $x_{k+1} = \prox{\frac{\tau}{\lambda}f}(z_k)$ \Comment*{Apply Proximity Operator}
  \If{$F(x_k) - F(x_{k+1}) \leq \frac{\gamma}{\tau} ||x_k - x_{k+1}||^2_2$ \Comment*{Backtracking}}{$\tau = \eta \tau$}
  \Else{$\Delta = \frac{F(x_k) - F(x_{k+1})}{F(x_0)}$\\
  $k = k + 1$ \Comment*{Go to next iteration if sufficient descent}}}
\Return{$\hat{x} = \tau D_\sigma(x_K) + (1 - \tau)x_K$ \Comment*{Discard residual noise}}
\label{alg:GS-PnP}
\end{algorithm}
\section{Proximal Plug-and-Play}\label{sec:prox-PnP}
We present the construction and training of the proximal denoiser in Section~\ref{sec:prox-denoiser}, and then detail the proximal plug-and-play algorithms in Section~\ref{sec:prox-pnp op}.
We recall convergence results in Section~\ref{sec:Prox-conv}. 
\subsection{Proximal Denoiser}\label{sec:prox-denoiser}
In this section we discuss another interpretation  of the GS-Denoiser that was presented in~\cite{ProxPnP}. Under some additional constraint, the GS-operator $D_\sigma$ can indeed be written as the proximity operator of an explicit function.
\begin{proposition}[Proximal denoisers from~\cite{ProxPnP}]\label{prop:prox-denoiser}
Let $g_\sigma : \mathbb{R}^n \rightarrow \mathbb{R}$ a $C^{k+1}$ function and $\nabla g_\sigma$ $ L_{g_\sigma}$-Lipschitz with $L_{g_\sigma} < 1$. Let $D_\sigma := \id - \nabla g_\sigma = \nabla h_\sigma$. Then,
\begin{enumerate}
    \item there exists a $\frac{L_{g_\sigma}}{L_{g_\sigma} + 1}$-weakly convex potential $\phi_\sigma : \mathbb{R}^n \rightarrow \mathbb{R} \cup \{+\infty\}$, such that Prox$_{\phi_\sigma}$ is one-to-one and
    \begin{equation}
        D_\sigma = \prox{\phi_\sigma}.
    \end{equation}
    Moreover $D_\sigma$ is injective, $\im(D_\sigma)$ is open and there is a constant $K \in \mathbb{R}$ such that $\phi_\sigma$ is defined on $\im(D_\sigma)$ by
    \begin{equation}
      \forall x \in \im(D_\sigma), \quad
      \phi_\sigma(x) = g_\sigma(D_\sigma^{-1}(x))) - \frac{1}{2}||D_\sigma^{-1}(x) - x||^2 + K . \\
    \label{phi}
    \end{equation}
    \item $\forall x \in \mathbb{R}^n, \phi_\sigma(x) \geq g_\sigma(x) + K$ and for $x \in \fix(D_\sigma), \phi_\sigma(x) = g_\sigma(x) + K$.
    \item $\phi_\sigma$ is $C^k$ on $\im(D_\sigma)$ and $\forall x \in $ $\im(D_\sigma), \nabla \phi_\sigma(x) = D_\sigma^{-1}(x) - x = \nabla g_\sigma (D_\sigma^{-1}(x))$. Moreover, $\nabla \phi_\sigma$ is $\frac{L_{g_\sigma}}{1 - L_{g_\sigma}}$-Lipschitz on $\im(D_\sigma)$.
\end{enumerate}
\end{proposition}
This result states the GS-Denoiser corresponds to the Proximal operator of the potential $\phi_\sigma$ as soon as $\nabla g_\sigma$ is a $L_g<1$ Lipschitz operator.
\paragraph{Training loss}\label{sec:train2}
Proposition \ref{prop:prox-denoiser} tells us that it is sufficient to ask for $\nabla g_\sigma$ to be a contraction so that the denoiser can effectively be a proximity operator.
To that aim, the approach of~\cite{spec-norm} is considered, that is, the Lipschitz constant of $\nabla g_\sigma$ is softly penalized by regularizing its Hessian matrix spectral norm $|||\nabla^2 g_\sigma||| = |||J_{(Id - D_\sigma)}|||$ computed in practice via power iterations .
The largest eigenvalue of a matrix will be approximated by using the power iteration algorithm recalled in Section~\ref{Prox-Denoiser}.

The proximal denoiser can be obtained as an adaptation of the GS-denoiser.
Training starts from the GS-denoiser weights obtained after training with loss (\ref{loss-gsd}), and then fine-tunes over several epochs using the augmented loss
\begin{align}\label{loss-prox}
    \mathcal{L}(D_\sigma) &= \mathbb{E}_{x \sim p_{data}, \xi_\sigma \sim \mathcal{N}(0, \sigma^2 I)}\left[ ||D_\sigma(x + \xi_\sigma) -  x||^2+ \mu \max( |||J_{(Id - D_\sigma)}(x + \xi_\sigma) |||, 1-\varepsilon)\right].
\end{align}
We notice that $\mu$ is a chosen parameter, in this work we use weights obtained with $\mu = 10^{-3}$.
\subsection{The Prox-PnP algorithm}\label{sec:prox-pnp op}
In this section we present the usage of the proximal denoiser explained in Section \ref{sec:prox-denoiser}.
Since we now consider the expression $D_\sigma = \prox{\phi_\sigma}$, the new objective function becomes
\begin{equation}\label{ref:Prox-energy}
    F^{\lambda, \sigma} = \frac{1}{\lambda} f + \phi_\sigma.
\end{equation}
We will then use classical proximal splitting algorithms such as Proximal Gradient Descent (PGD) and Douglas-Rashford Splitting (DRS), in the Plug-and-Play fashion. In~\cite{ProxPnP}, these algorithms are called  Prox-PnP algorithms. We can write the considered  operators for $\beta \in (0, 1]$:
\begin{align}
  \textbf{(PGD) } T^{\tau, \lambda, \sigma}_{PGD}
  &= \prox{\tau \phi_\sigma} \circ \left(\id - \frac{\tau}{\lambda} \nabla f\right)\\
  \textbf{(DRS) } T^{\tau, \lambda, \sigma}_{DRS}
  &= \beta \left(\mathsf{RProx}_{\tau \phi_\sigma} \circ \mathsf{RProx}_{\frac{\tau}{\lambda} f}\right) +(1 - \beta) \id
\end{align}
where $\mathsf{RProx}_f$ is the reflected proximity operator of $f$, defined by
\begin{equation}
    \mathsf{RProx}_f := 2 \prox{f} - \id.
\end{equation}

We need to separate the cases where the data-fidelity term $f$ is differentiable or not. When $f$ is differentiable we consider the PGD algorithm and a DRSdiff algorithm that are simply induced from the classical versions when plugging in the proximal denoiser. For the non-differentiable case, since $\phi_\sigma$ is differentiable we swap their roles in the algorithm, this also gets rid of the possible restrictions on the regularization parameter $\lambda$. 
We also have to set $\tau =1$, since we only have access to  $\prox{\phi_\sigma}$ and not to $\prox{\tau\phi_{\sigma}}$, for any $\tau>0$. We finally obtain
\begin{align}
    \textbf{(Prox-PnP-PGD) } T^{\lambda, \sigma}_{Prox-PnP-PGD} &= D_\sigma \circ \left(\id - \frac{1}{\lambda} \nabla f\right)\label{op:Prox-PGD}\\
    \begin{split}\label{op:Prox-DRSdiff}
        \textbf{(Prox-PnP-DRSdiff) } T^{\lambda, \sigma}_{Prox-PnP-DRSdiff} &= \beta \left(2D_\sigma \circ \mathsf{RProx}_{\frac{1}{\lambda} f} - \mathsf{RProx}_{\frac{1}{\lambda} f}\right) +(1 - \beta) \id.\\
        &= \id + 2 \beta \left(D_\sigma \circ \mathsf{RProx}_{\frac{1}{\lambda} f} - \prox{\frac{1}{\lambda} f}\right)
    \end{split}\\
        \textbf{(Prox-PnP-DRS) } T^{\lambda, \sigma}_{Prox-PnP-DRS}
        &= \id + 2 \beta \left(\mathsf{Prox}_{\frac{1}{\lambda} f} \circ \left (2 D_\sigma - \id \right)  - D_\sigma\right)\label{op:Prox-DRS}
\end{align}

We then define the associated fixed point iteration algorithms as for the GS-PnP scheme~\eqref{GSPnP-Scheme}.
\subsection{Convergence analysis}\label{sec:Prox-conv}
In this section we recall convergence results for each of the fixed point iteration schemes associated to the operators described above.
These convergence results rely on a Lyapunov function associated to an optimization scheme over several variables, from which we can deduce convergence of the iterates towards a stationary point of the objective function \eqref{ref:Prox-energy}, which corresponds to a fix point of the iterated operator.

First let us share a theorem that proves the convergence of the fixed point iteration scheme given by operator \eqref{op:Prox-PGD}: $x_0 \in \mathbb{R}^n, \forall k \geq 0, x_{k+1} = T^{\lambda, \sigma}_{Prox-PnP-DRS}(x_k)$.
\newpage

\begin{theorem}[Convergence of Prox-PnP-PGD, Theorem 4.1 in~\cite{ProxPnP}]\label{Thm: PGD}
    Let $g_\sigma : \mathbb{R}^n \rightarrow \mathbb{R} \cup \{+\infty\}$ of class $C^2$ with $L$-Lipschitz gradient, $L < 1$, and $D_\sigma := \id - \nabla g_\sigma$. Let $\phi_\sigma$ defined from $g_\sigma$ and $D_\sigma$ as in (\ref{phi}). Let $f : \mathbb{R}^n \rightarrow \mathbb{R} \cup \{+\infty\}$ differentiable with $L_f$ -Lipschitz gradient. Assume that $f$ and $g_\sigma$ are bounded from below. Then, for $ L_f < \lambda$, the iterates $x_k$ given by the iterations (\ref{op:Prox-PGD}) verify
\begin{enumerate}
    \item $(F^{\lambda,\sigma}(x_k))$ is non-increasing and converges.
    \item The residual $||x_{k+1} - x_k||$ converges to $0$ at rate
$\min_{k \leq K} ||x_{k+1} - x_k|| = O(\frac{1}{\sqrt{K}})$.
    \item All cluster points of the sequence $(x_k)$ are stationary
points of $F^{\lambda,\sigma}$.
    \item Additionally suppose that $f$ and $g_\sigma$ are respectively KŁ and semi-algebraic, then if the sequence $(x_k)$ is bounded, it converges, with finite length, to a stationary point of $F^{\lambda,\sigma}$.
\end{enumerate}
\end{theorem}

For Douglas-Rachford plug-and-play, two variants will be given.
The first one is given by the iterations of operator~\eqref{op:Prox-DRSdiff}, which can be decomposed as
\begin{align}\label{Scheme-DRSdiff}
   \begin{cases}
    u_{k+1}=\prox{\frac{1}{\lambda} f}(x_k)\\
    v_{k+1} = D_\sigma(2u_{k+1} - x_k) = \prox{\phi_\sigma} (2u_{k+1} - x_k) \\
    x_{k+1} = x_k + (v_{k+1} - u_{k+1}).
    \end{cases} 
\end{align}
The corresponding Lyapunov function writes
\begin{align}
    F^{DR,1}_{\sigma,\lambda}(x) = \phi_\sigma(v) + \frac{1}{\lambda} f(u) + \langle u-x,u-v\rangle+ \frac{1}{2}||u - v||^2 ,
\end{align}
where for a fixed $x$, we compute $u$ and $v$ from the operations given in equations (\ref{Scheme-DRSdiff}).
\begin{theorem}[Convergence of Prox-PnP-DRSdiff, Theorem 4.3 in~\cite{ProxPnP}]\label{Thm: DRSdiff}
    Let $g_\sigma : \mathbb{R}^n \rightarrow \mathbb{R} \cup \{+\infty\}$ of class $C^2$ with $L$-Lipschitz gradient, $L < 1$, and $D_\sigma := \id - \nabla g_\sigma$. Let $f : \mathbb{R}^n \rightarrow \mathbb{R} \cup \{+\infty\}$ be convex and differentiable with $L_f$ -Lipschitz gradient. Assume that $f$ and $g_\sigma$ are bounded from below. Then, for $ L_f < \lambda$, the iterates $(u_k , v_k , x_k )$ given by the iterative scheme (\ref{Scheme-DRSdiff}) verify
\begin{enumerate}
    \item $(F^{DR,1}(x_k))$ is non-increasing and converges.
    \item The residual $||u_k - v_k||$ converges to $0$ at rate $\min_{k \leq K} ||u_{k} - v_k|| = O(\frac{1}{\sqrt{K}})$.
    \item $(u_k)$ and $(v_k)$ have the same cluster points, all of them being stationary for $F^{\lambda,\sigma}$ with the same value of $F^{\lambda,\sigma}$.
    \item Additionally suppose that $f$ and $g_\sigma$ are respectively KŁ and semi-algebraic, then if the sequence $(u_k , v_k , x_k )$ is bounded, the whole sequence converges, and $(u_k)$ and $(v_k)$ converge to the same stationary point of $F^{\lambda,\sigma}$.
\end{enumerate}
\end{theorem}

The second plug-and-play Douglas-Rachford scheme is obtained through the iterations of the operator \eqref{op:Prox-DRS}, which can be decomposed as
\begin{align}\label{Scheme-DRS}
   \begin{cases}
    u_{k+1}= D_\sigma(x_k) = \prox{\phi_\sigma}(x_k)\\
    v_{k+1} = \prox{\frac{1}{\lambda} f}(2u_{k+1} - x_k)  \\
    x_{k+1} = x_k + (v_{k+1} - u_{k+1})
    \end{cases} 
\end{align}
for which we adapt the Douglas-Rachford Envelope:
\begin{align}
    F^{DR,2}_{\sigma,\lambda}(x) = \phi_\sigma(u) + \frac{1}{\lambda} f(v) + \langle u-x,u-v \rangle+ \frac{1}{2}||u - v||^2 .
\end{align}

Contrarily to the two previous theorems, the next result gives convergence of Prox-PnP-DRS without any constraint on the regularization term $\lambda$. However, this theorem involves a more restrictive Lipschitz constraint $L<1/2$ on the denoiser, which limits its expressive power and thus its performance. Finally note that~\cite{renaud2025stability}[Appendix D, Proposition 2] ensures that $\im(D_\sigma)$ is almost convex since as $\phi_\sigma$ is weakly convex (Proposition~\ref{prop:prox-denoiser}).

\begin{theorem}[(Convergence of Prox-PnP-DRS with non-differentiable $f$, Theorem 4.4 in~\cite{ProxPnP}]\label{Thm: DRS}
    Let $g_\sigma : \mathbb{R}^n \rightarrow \mathbb{R} \cup \{+\infty\}$ of class $C^2$ with $L$-Lipschitz gradient, $L < \frac{1}{2}$, and $D_\sigma := \id - \nabla g_\sigma$. Assume that $\im(D_\sigma)$ is convex. Let $f : \mathbb{R}^n \rightarrow \mathbb{R} \cup \{+\infty\}$ be proper lower-semicontinuous. Assume that $f$ and $g_\sigma$ are bounded from below. Then, $\forall \lambda > 0$, the iterates $(u_k , v_k , x_k)$ given by the scheme (\ref{Scheme-DRS}) verify
    \begin{enumerate}
        \item $(F^{DR,2}_{\lambda, \sigma}(x_k))$ is non-increasing and converging.
        \item The residual $||u_k - v_k||$ converges to $0$ at rate $\min_{k \leq K} ||u_{k} - v_k|| = O(\frac{1}{\sqrt{K}})$.
        \item For any cluster point $(u^*, v^*, x^*)$, $u^*$ and $v^*$ coincides to a stationary point of $F^{\lambda,\sigma}$ .
        \item Also suppose that $f$ and $g_\sigma$ are respectively KŁ and semi-algebraic, then if the sequence $(u_k , v_k , x_k )$ is bounded, it  converges, and $(u_k)$ and $(v_k)$ converge to the same stationary point of $F^{\lambda,\sigma}$.
    \end{enumerate}
\end{theorem}
For DRS algorithms, looking at the above theorems we understand that $(u_k)$ and $(v_k)$ have the same limit when $k \rightarrow +\infty$, which is a stationary point of the objective \eqref{ref:Prox-energy}. That said since we fix a finite number of iterations it is good practice to consider as final term the output of the denoiser during the final iteration, to remove residual noise. We make sure of this practice in the following algorithm pseudo-codes given in Algorithms \ref{alg:Prox-PGD}, \ref{alg:Prox-DRSdiff} and~\ref{alg:Prox-DRS}.\vspace{-.1cm}
\begin{algorithm}[!ht]
\caption{Prox-PnP-PGD image restoration}
\DontPrintSemicolon
\Param{init. $z_0, \lambda > L_f, \sigma > 0, \epsilon > 0, \tau_0 > 0, K \in \mathbb{N}^*, \eta \in (0, 1), \gamma \in (0, 1/2)$}
\Input{degraded image $y$}
\Output{restored image $\hat{x}$}
$k = 0$\\
\While{$k < K$}{
  $z_k = x_k - \frac{1}{\lambda} \nabla f(x_k)$ \Comment*{Apply Gradient Descent on data fidelity term}
  $x_{k+1} =D_\sigma(z_k)$ \Comment*{Use Prox-Denoiser to apply Proximal Step}
  $k = k + 1$}
\Return{$\hat{x} = x_K$ \Comment*{Return last denoised iteration}}
\label{alg:Prox-PGD}
\end{algorithm}
\begin{algorithm}[!ht]
\caption{Prox-PnP-DRSdiff image restoration}
\DontPrintSemicolon
\Param{init. $x_0,\lambda > L_f, \sigma > 0, \epsilon > 0, \beta \in (0,1], K \in \mathbb{N}^*$}
\Input{degraded image $y$}
\Output{restored image $\hat{x}$}
$k = 0$\\
\While{$k < K$}{
  $u_k = \prox{\frac{1}{\lambda} f}(x_k)$ \Comment*{Apply Proximity operator}
  $v_{k} = D_\sigma(2 u_k - x_k)$ \Comment*{Use Prox-Denoiser to apply Proximal Step on Reflexive-prox}
  $x_{k+1} =  x_k + 2\beta ( v_k - u_k)$ \Comment*{Apply $\beta$-averaging}
  $k = k + 1$}
\Return{$\hat{x} = v_K$ \Comment*{Return last denoised iteration}}
\label{alg:Prox-DRSdiff}
\end{algorithm}
\begin{algorithm}[!ht]
\caption{Prox-PnP-DRS image restoration}
\DontPrintSemicolon
\Param{init. $z_0,\lambda > 0, \sigma > 0, \epsilon > 0, \beta \in (0,1], K \in \mathbb{N}^*$}
\Input{degraded image $y$}
\Output{restored image $\hat{x}$}
$k = 0, x_0 = \prox{\tau f}(z_0)$\\
\While{$k < K$}{
  $u_{k+1} = D_\sigma(x_k)$ \Comment*{Use Prox-Denoiser to apply Proximal Step}
  $v_{k+1} = \prox{\frac{1}{\lambda} f}(2 u_{k+1} - x_k)$ \Comment*{Apply Proximity operator on Reflexive-prox}
  $x_{k+1} =  x_k + 2\beta ( v_{k+1} - u_{k+1})$ \Comment*{Apply $\beta$-averaging}
  $k = k + 1$}
\Return{$\hat{x} = u_K$ \Comment*{Return last denoised iteration}}
\label{alg:Prox-DRS}
\end{algorithm}
\section{Training} \label{sec:Training}
In this section we provide technical and practical information for implementation. In Section~\ref{ssec:DRUNET} we present the DRUNet neural network architecture that is used to parameterize the GS-Denoisers, and in Section~\ref{ssec:WiDiP} we discuss how the training losses are computed and share details on the used dataset.
\subsection{The DRUNet Architecture}\label{ssec:DRUNET}
\begin{figure}[!ht]
    \centering
\includegraphics[width=\textwidth]{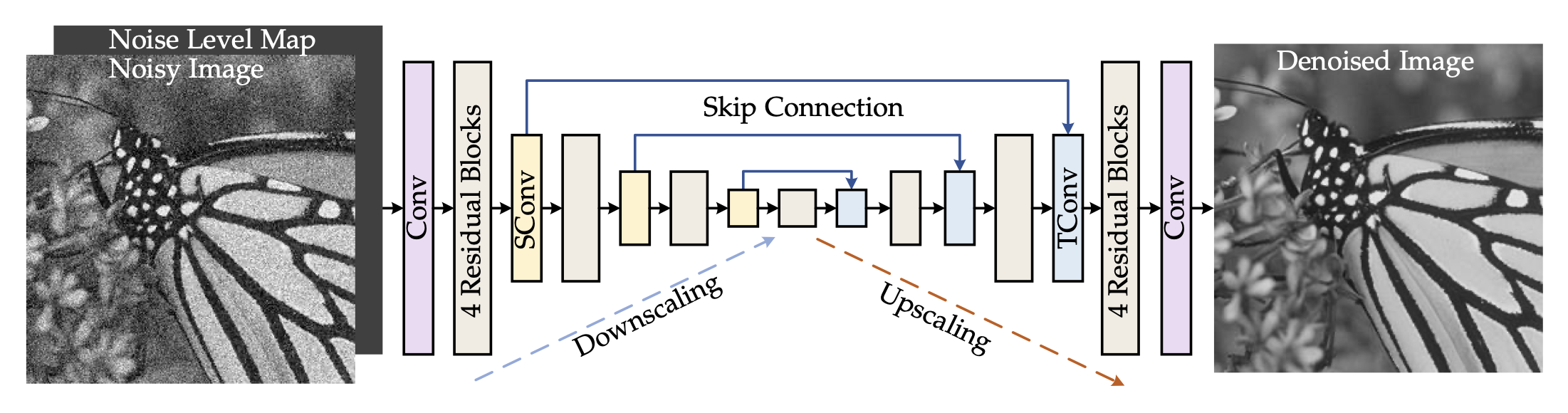}
    \caption{Architecture of the DRUNet denoiser~\cite{DRUnet}. Original image from their work.}
    \label{fig:Drunet}
  \end{figure}

The choice of parameterization for $N_\sigma$ is free and it seems natural to rely on a state-of-the-art architectures used for denoising.
In~\cite{hurault2021gradient} and~\cite{ProxPnP}, a DRUnet architecture is chosen. This architecture, presented in~\cite{DRUnet}, was constructed specifically for Plug-and-Play restoration. This model is built as a sequence of (strided) convolutions (and transposed convolutions) and ResNet blocks~\cite{resnet} placed in a UNet structure~\cite{Unet}. It takes as input a noisy image and a noise level map which corresponds to a constant grid containing the image noise level variance~$\sigma^2$. This noise level map is of same size as the input image and constitutes an additional channel, such that for images of shape $M \times N \times C$ the denoiser takes as input a tensor of shape $M \times N \times (C+1)$. Figure~\ref{fig:Drunet} illustrates this architecture.

As it is explained in the Remark~\ref{rem:lipschitz}, we need the denoiser to be differentiable. A neural network is a succession of linear transformations and activation functions. We need to choose activation functions that conserve differentiability, for that  the Exponential Linear Unit (ELU) activation function is used in~\cite{hurault2021gradient}. This activation function is a smooth version of the Rectified Linear Unit (ReLU) and has the expression for some $\alpha > 0$ (we note that in practice we fix $\alpha = 1$ to make it $C^1$):
\begin{equation} 
    ELU(x) :=
    \begin{cases}
    x &\text{if $x > 0$}\\
    \alpha(\exp(x) - 1) &\text{otherwise}.
    \end{cases}
\end{equation}

\subsection{Training Details}\label{ssec:WiDiP}
\subsubsection{Batch training}
In Section~\ref{sec:train1} we have shared the theoretical expression of the training loss.
In practice we use Monte-Carlo estimation to approximate the training loss with samples taken in a random batch of data points:
\begin{equation}
    \tilde{L} = \frac{1}{B}\sum_{i = 0}^B ||\nabla g_\sigma(x_i + \xi_\sigma) -  \xi_\sigma||^2 \approx \mathcal{L}(D_\sigma),
\end{equation}
where $B$ corresponds to the batch size.
The noise standard deviation $\sigma$ is randomly chosen in $[0, 50/255]$, but the noise level map is the same for each image in a batch. 

Given the dataset of colored natural images, described below in Section~\ref{dataset},  patches of size $128 \times 128$ are  randomly sampled as training set. They used a batch size of 16, and trained during 1500 epochs with the ADAM optimizer and a learning rate of $10^{-4}$ divided by 2 every 300 epochs.

\subsubsection{Proximal Denoiser}\label{Prox-Denoiser}

For the Prox-DRUNet, since convergence of its associated algorithms relies on $g_\sigma$ being of class $C^2$ (see (\ref{Thm: PGD}), (\ref{Thm: DRSdiff}) and (\ref{Thm: DRSdiff})), a Softplus activation function  of class $C^\infty$ is used in~\cite{ProxPnP}, in place of the ELU that is at best of class $C^1$ for $\alpha = 1$. The Softplus function is defined for a parameter $\beta > 0$ by
\begin{equation}
    \mathsf{Softplus}(x) := \frac{1}{\beta} \log (1+\exp(\beta x)),
\end{equation}
In practice, a Pytorch implementation and the default value $\beta = 1$ are used in~\cite{ProxPnP}. 

Again we need to approximate the expectations that define the training loss (\ref{loss-prox}) for the Prox-denoiser. This also relies on Monte-Carlo estimation. Additionally we must estimate  the  spectral norm of the Hessian matrix of $g_\sigma$, which is equal to the Prox-Denoiser Jacobian matrix largest singular value that is approximated using the power iteration algorithm, given in  Algorithm~\ref{alg:pow_it}.
In practice, 50 iterations of power iteration are realized for each batch during training. The denoiser was fine-trained for different values of the penalization parameter $\mu$ (see loss \eqref{loss-prox}) during 10 epochs.

\begin{algorithm}[!ht]
\caption{Power iteration algorithm}
\DontPrintSemicolon
\Param{init. $b_0, K \in \mathbb{N}^*$}
\Input{diagonalisable matrix of interest $A$ of size $n \times m$}
\Output{largest singular value of $A$}
$k = 0, b_0 =$ random vector of size $m$\\
\While{$k < K$}{
    $b_k^1 = A b_k$\\
    $b_{k+1} = \frac{b_k^1}{||b_k^1||}$
}
\Return{$\frac{b_K^T A b_K}{||b_K||^2}$\Comment*{return Rayleigh quotient}}
\label{alg:pow_it}
\end{algorithm}

\subsubsection{Dataset}\label{dataset}
The same dataset that was used to train the original DRUNet~\cite{DRUnet} is considered, which is a combination of the Berkeley segmentation dataset (CBSD)~\cite{CBSD}, Waterloo Exploration Database~\cite{WED}, DIV2K dataset~\cite{DIV2K} and Flick2K dataset~\cite{FLICK2K}. These datasets contain colored natural images.

\section{Experiments}\label{sec:Exp}
In this section, we provide numerical experiments which, on one side, assess the performance of the introduced denoisers (Section~\ref{ssec:res:denoising}), and on the other side, examine the behavior of the presented plug-and-play algorithms on three different inverse problems, with a particular focus on the impact of the main parameters $\sigma$ and $\lambda$.

\subsection{Denoising}\label{ssec:res:denoising}

First, let us evaluate the performance of the gradient-step and proximal denoisers. We will rely on the peak signal-to-noise ratio (PSNR).
Looking at the results in Figure~\ref{fig:denoising} and Figure~\ref{fig:psnr_over_sigma}, we notice that the GS-Denoiser has slightly better performance than the Prox-Denoiser when applied on image degraded by noise levels within the training range $[0,50]/255$. This comes from the fact that the Prox-Denoiser has an additional constraint during training that keeps it from having better denoising capabilites. But as a counterpart the standard deviations (due to the random input noise) are smaller with the Prox-Denoiser. Moreover, the constraint on the Lipschitz constant of $\nabla g_\sigma$ in the expression of $D_\sigma = \text{Id }- \nabla g_\sigma$ seems to induce more robustness to variations since it better generalizes to unseen noise levels up to $50 / 255$ (see second row of Figure~\ref{fig:denoising}). Some artifacts  appear for the GS-denoiser when the noise level is high, which may be caused by over-fitting during training. 
This over-fitting is reduced when regularizing with respect to the Lipschitz coefficient in the Prox-PNP framework, which is confirmed by Figure~\ref{fig:psnr_over_sigma}.

\newcommand{\sidecapX}[1]{ {\begin{sideways}\parbox{0.22\textwidth}{\centering #1}\end{sideways}} }
 \setlength{\fboxsep}{0pt}
\begin{figure}[!ht]
\begin{center}
\begin{tabular}{cccc}
&Observation&GS denoiser &Prox denoiser\\
\sidecapX{$\sigma = 72 / 255$}&\includegraphics[width=0.22\textwidth]{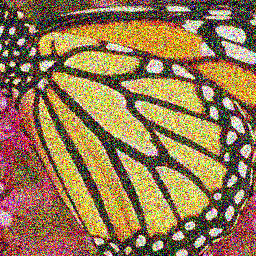} &\includegraphics[width=0.22\textwidth]{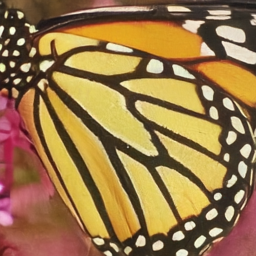} &
        \includegraphics[width=0.22\textwidth]{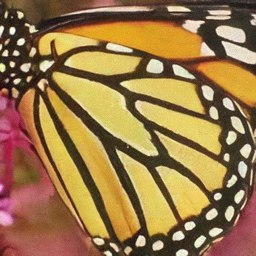} \\
        && \textbf{$\psnr=26.69 \pm 0.08$} & \textbf{$\psnr=25.74 \pm 0.06$}\\
\sidecapX{$\sigma = 125 / 255$}&\includegraphics[width=0.22\textwidth]{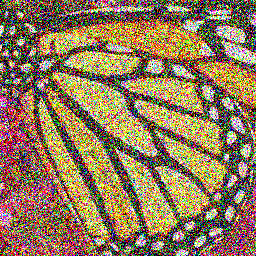}& 
        \includegraphics[width=0.22\textwidth]{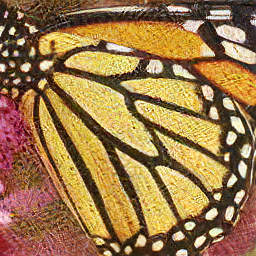}&  
        \includegraphics[width=0.22\textwidth]{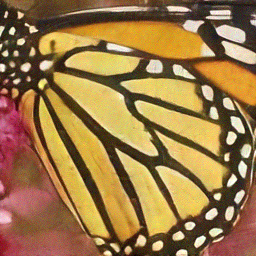} 
\\
&& \textbf{$\psnr=19.36 \pm 0.14$} & \textbf{$\psnr=22.57 \pm 0.05$}
\end{tabular} 
\end{center}
    \caption{Denoising results for different noise levels with the GS-DRUNet architecture trained as the GS-denoiser (second column) or the Prox-denoiser (third column). Denoising was run 10 times with different noise realizations for each image, and we share the mean $\psnr \pm$ standard-deviation.}
    \label{fig:denoising}
\end{figure}

\begin{figure}[!ht]
    \centering
    \includegraphics[width=0.5\textwidth]{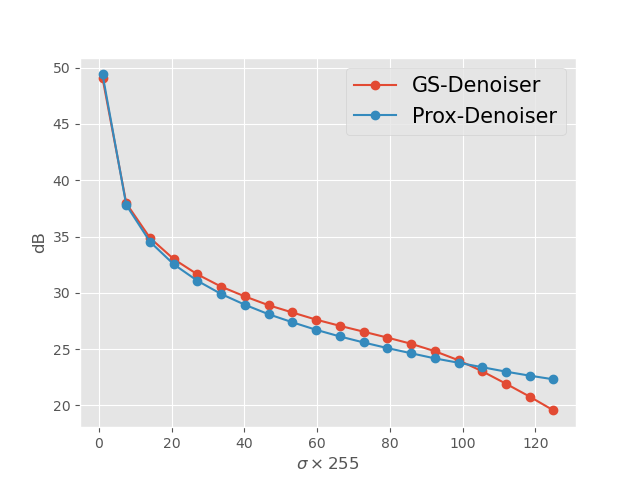}\\
    \caption{Mean denoising $\psnr$ evolution w.r.t $\sigma$. We average PSNR results over the images of set3C.
      We see that for low noise levels both denoisers have similar performance.
      For larger noise levels, the GS-denoiser is slightly more efficient, except for noise level highly out of the range used for training ($\sigma \geq 100 / 255$), which seems to indicate that the Prox-denoiser generalizes better.}
    \label{fig:psnr_over_sigma}
\end{figure}

\subsection{Plug-and-Play Image restoration} \label{ssec:restoration}
We now provide experiments for different image restoration tasks (super-resolution, deblurring and random inpainting) using the GS-PnP algorithm and the Prox-PnP algorithms. We explicit the impact of each parameter on the obtained results. For this we start from a configuration of parameters advised in~\cite{hurault2021gradient} and~\cite{ProxPnP}, and we tweak each parameter on its own to study its impact. 
\begin{remark}
    Each problem has its own associated linear operator, which can make the data-fidelity term differentiable and / or proximable. We will make that explicit for each tackled problem.
\end{remark} 
\paragraph{Super-resolution}
Super-resolution consists in starting from a low resolution image of size $n \times m$ and trying to obtain a high resolution version of this image of size $ns \times ms$, for a scaling factor $s>1$ (e.g. $s = 2$, $3$ or $4$).
We thus have $x \in \mathbb{R}^N$  and $y\in \mathbb{R}^M$ with $M = nm$ and $N = Ms^2$.

We then need to compare the different high-resolution propositions of $x$ with the low-resolution version $y $ by considering the operator $A$ as a composition of two steps:
$i)$ apply an anti-aliasing filter~$H$,
$ii)$ down-sample the image with a $s$-fold down-sampling matrix $D$.
So the considered data-fidelity term can be written (in the additive Gaussian noise model) as $f(x) = \frac{1}{2}||DHx - y||^2_2$ and the closed-form solution to its proximity operator is given in~\cite{prox-sr}:
\begin{equation}\label{sr-Prox}
    \prox{\tau f}(z) = \hat{z}_\tau -\frac{1}{s^2} \mathcal{F}^* \Lambda^* (I_M + \frac{\tau}{s^2}\Lambda \Lambda^*)^{-1}\Lambda \mathcal{F} \hat{z}_\tau,
\end{equation}
where $\hat{z}_\tau = \tau H^T D^T y + z$, $\mathcal{F}$ is the orthogonal matrix associated to the Fourier transform and finally 
${\Lambda = [\Lambda_1,...,\Lambda_{s^2}] \in \mathbb{R}^{M \times N}}$,
with $\underline{\Lambda}  = \text{diag}(\Lambda_1,...,\Lambda_{s^2})$, is a block-diagonal decomposition according to a $s \times s$ paving of the Fourier domain such that $H = \mathcal{F}^* \underline{\Lambda} \mathcal{F}$.
Notice that $I_M + \frac{\tau}{s^2}\Lambda \Lambda^*$ is a $M \times M$ diagonal matrix that can be easily inverted.

\begin{figure}[!ht]
    \centering
    \resizebox{\textwidth}{!}{
    \begin{tabular}{c c c c c c c c}
     \includegraphics[width=0.100\textwidth]{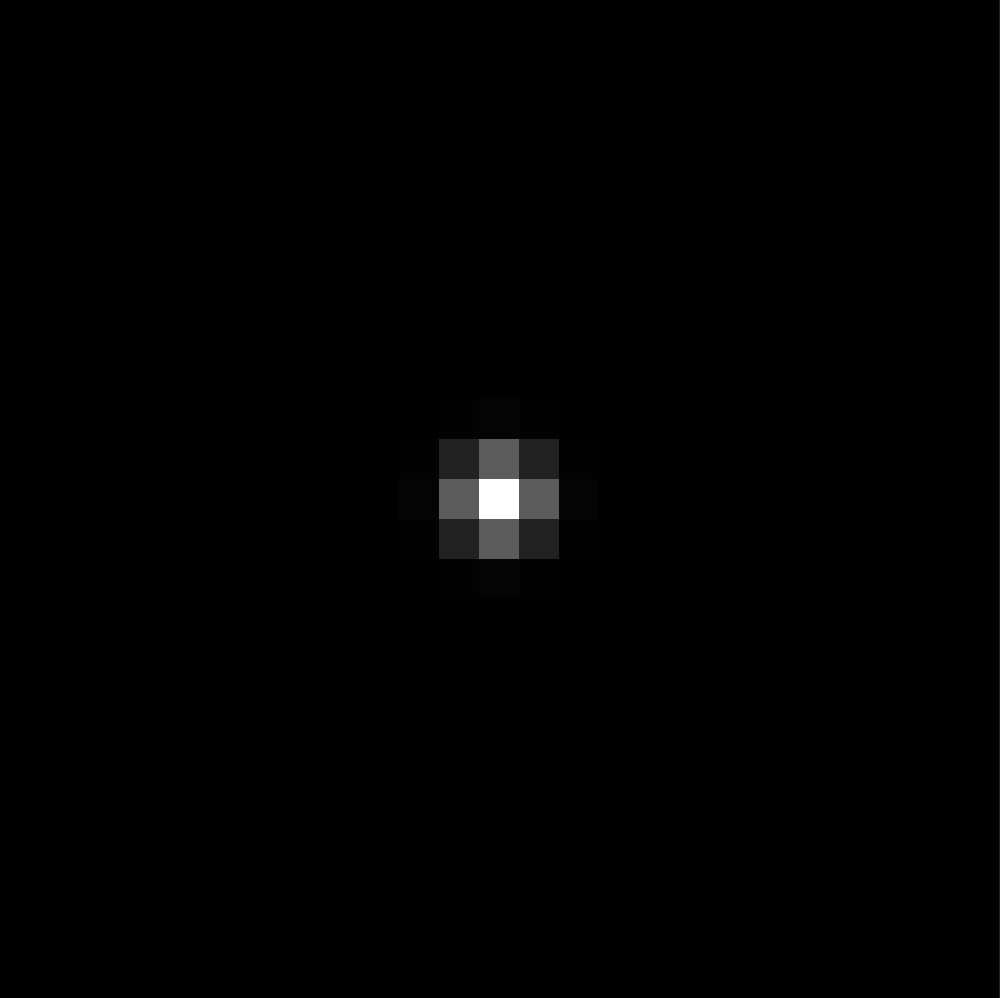} & 
      \includegraphics[width=0.100\textwidth]{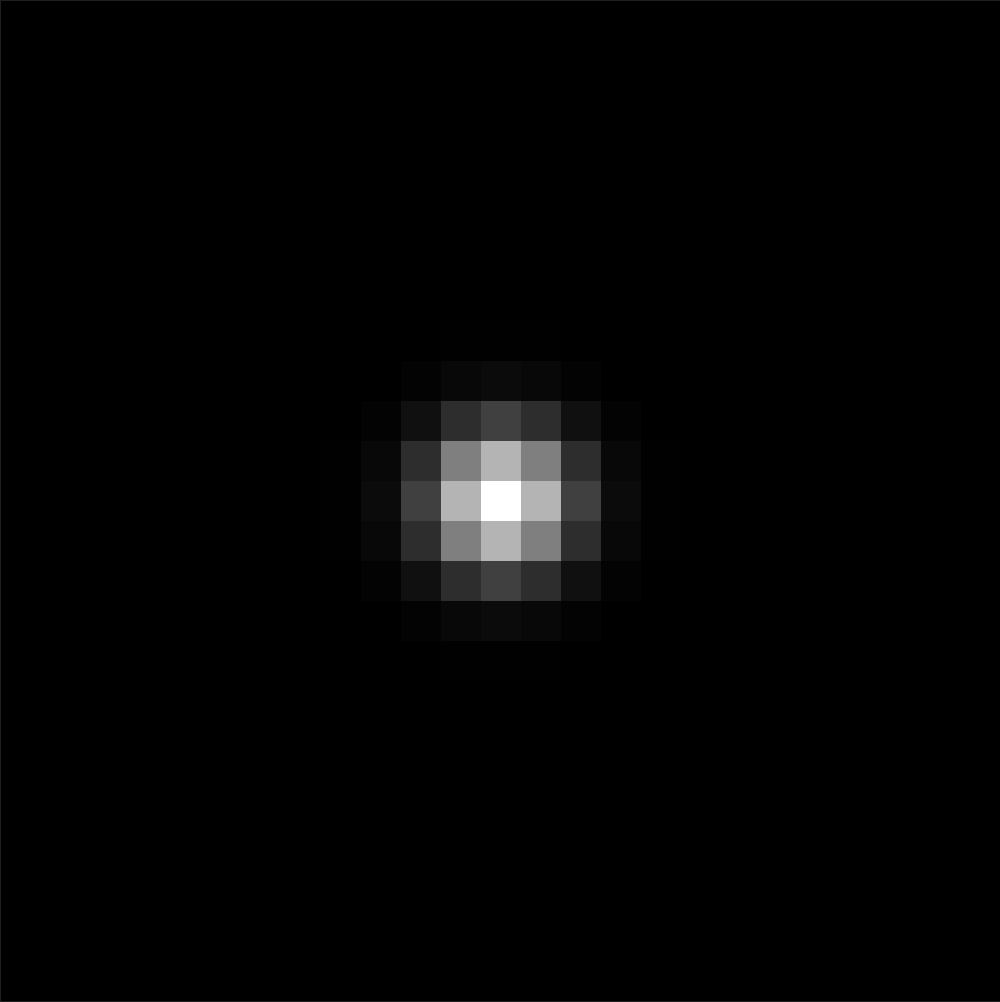} & 
       \includegraphics[width=0.100\textwidth]{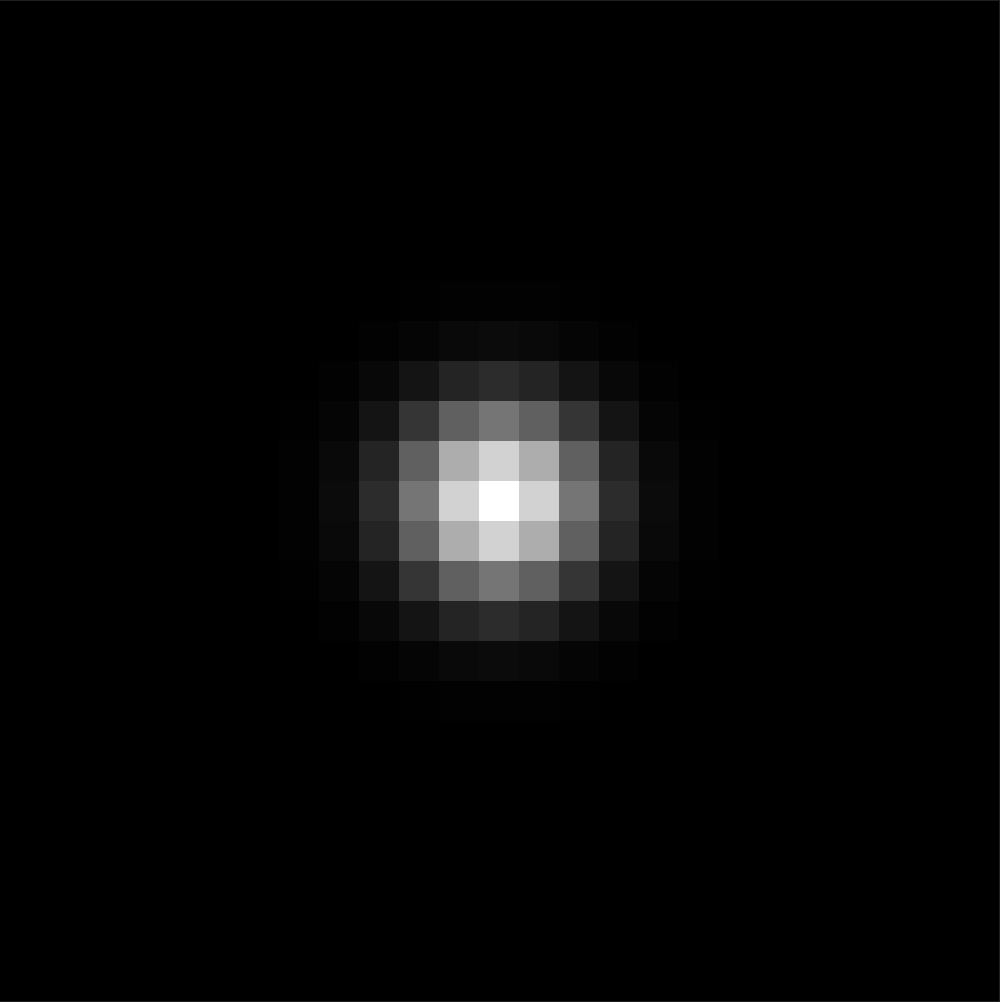} & 
        \includegraphics[width=0.100\textwidth]{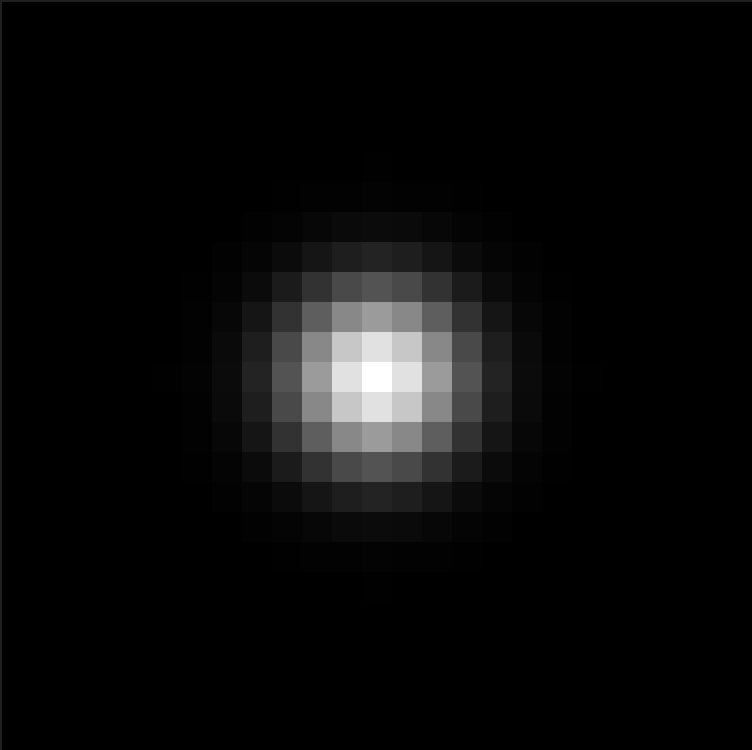}& 
         \includegraphics[width=0.100\textwidth]{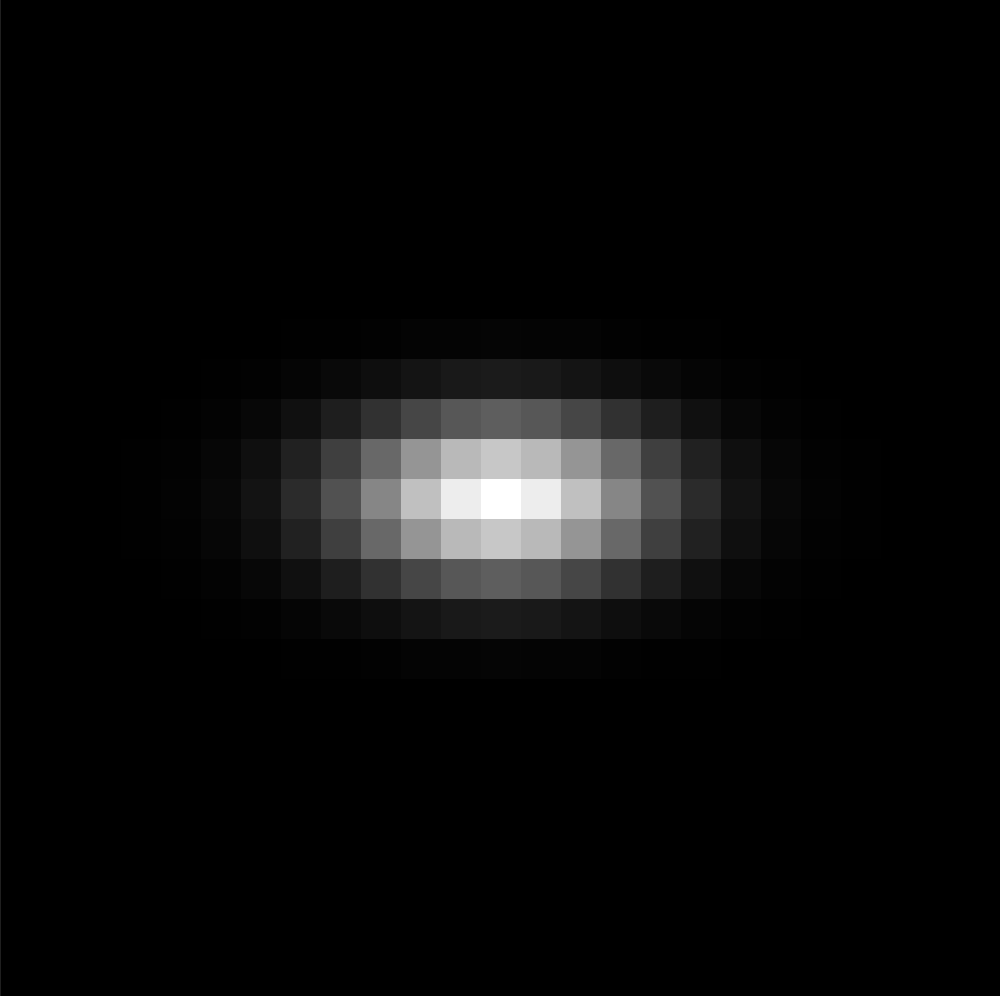} & 
      \includegraphics[width=0.100\textwidth]{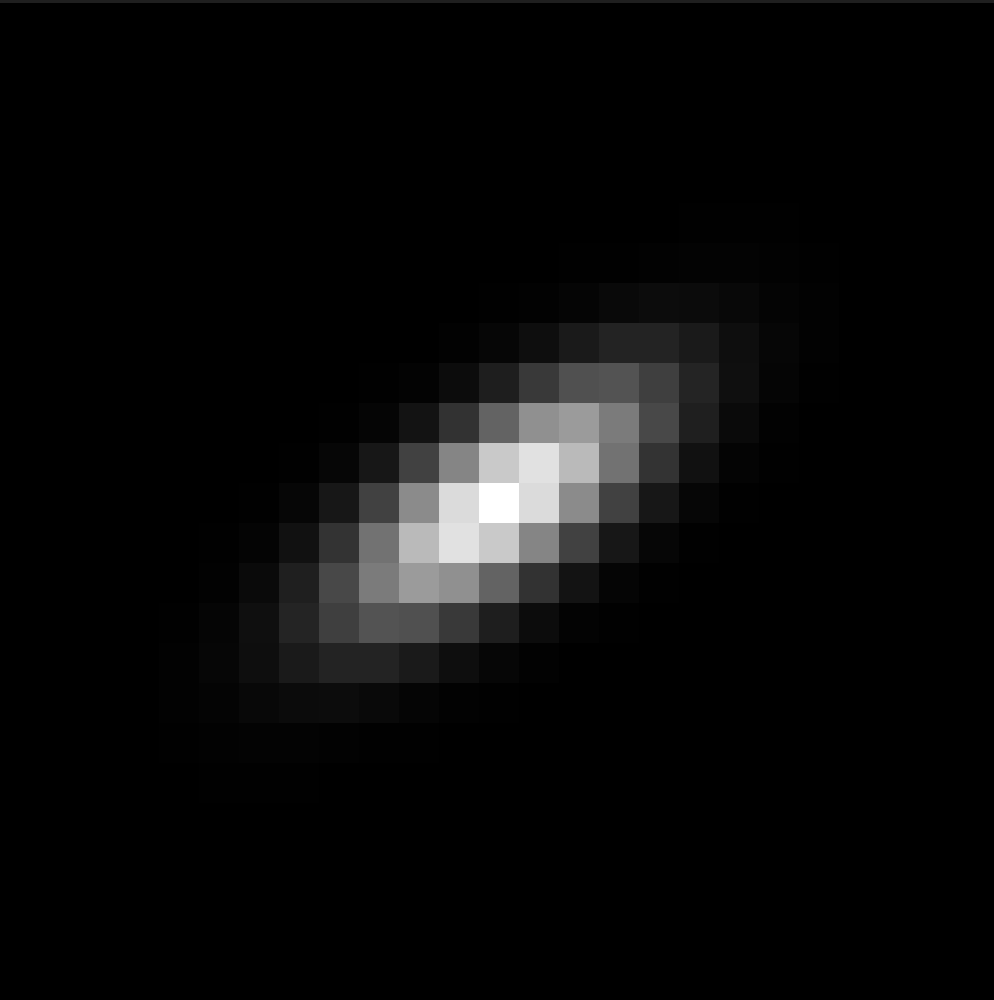} & 
       \includegraphics[width=0.100\textwidth]{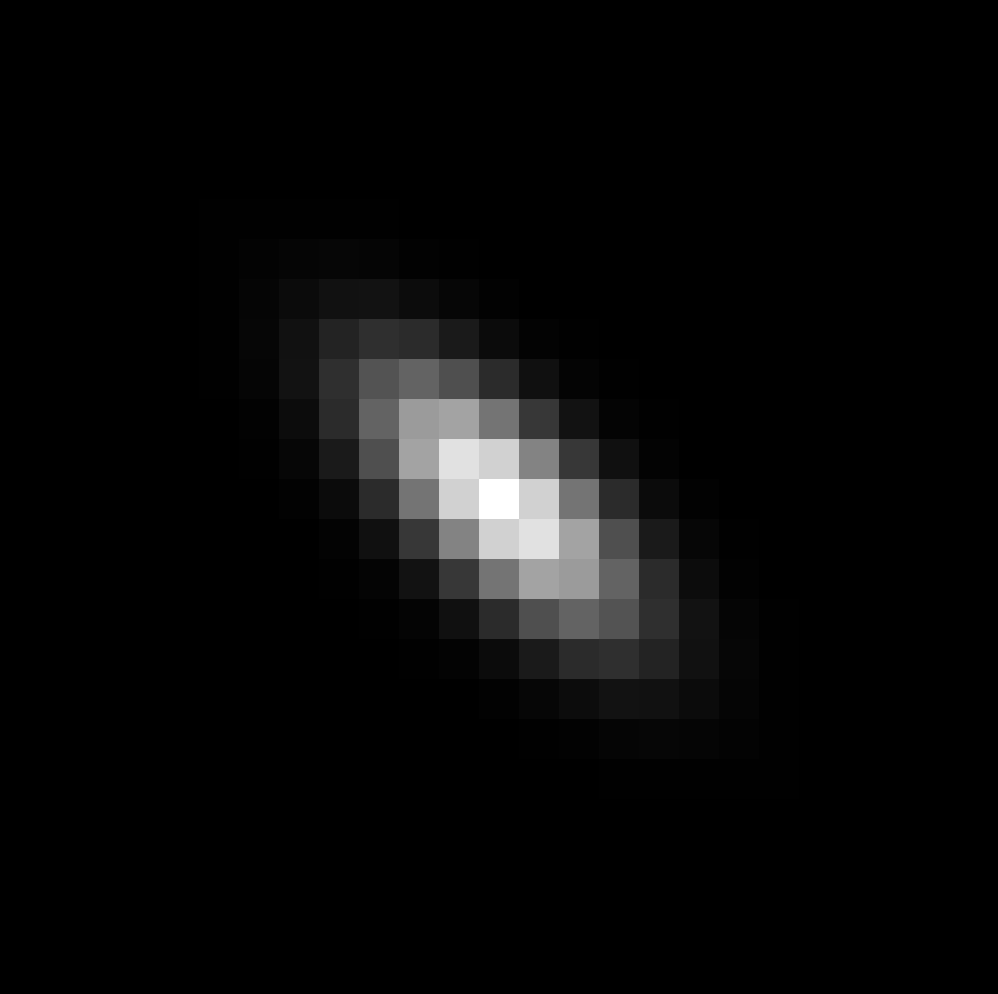} & 
        \includegraphics[width=0.100\textwidth]{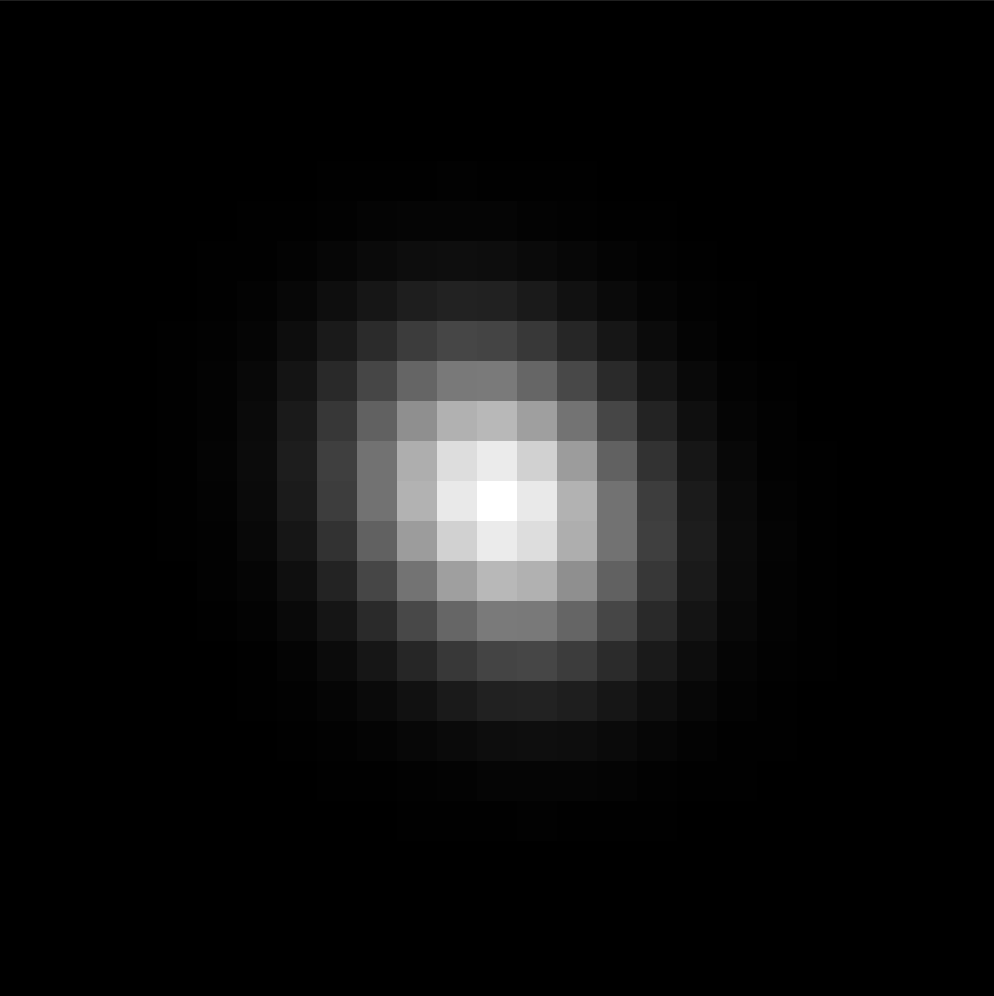}\vspace{-0.1cm}\\
    \end{tabular}
    }
    \caption{Filters used for the super-resolution problem.}
    \label{fig:aa-filters}
\end{figure}

Authors of~\cite{DRUnet} mention that when interpolating the low-resolution image to obtain an initial high-resolution guess, the pixels will tend to shift to incorrect positions (compared to the ground truth). To correct this shift they use a 2D linear grid interpolation to shift the pixels in the correct positions. We use bicubic interpolation to have an initial proposal for super-resolution. As in~\cite{RED}, we use standard isotopic and non-isotopic Gaussian anti-aliasing filters that are shown in Figure~\ref{fig:aa-filters}.

\begin{figure}[!ht]
    \centering
    \resizebox{\textwidth}{!}{
    \begin{tabular}{c c c c c c c c}
     \includegraphics[width=0.100\textwidth]{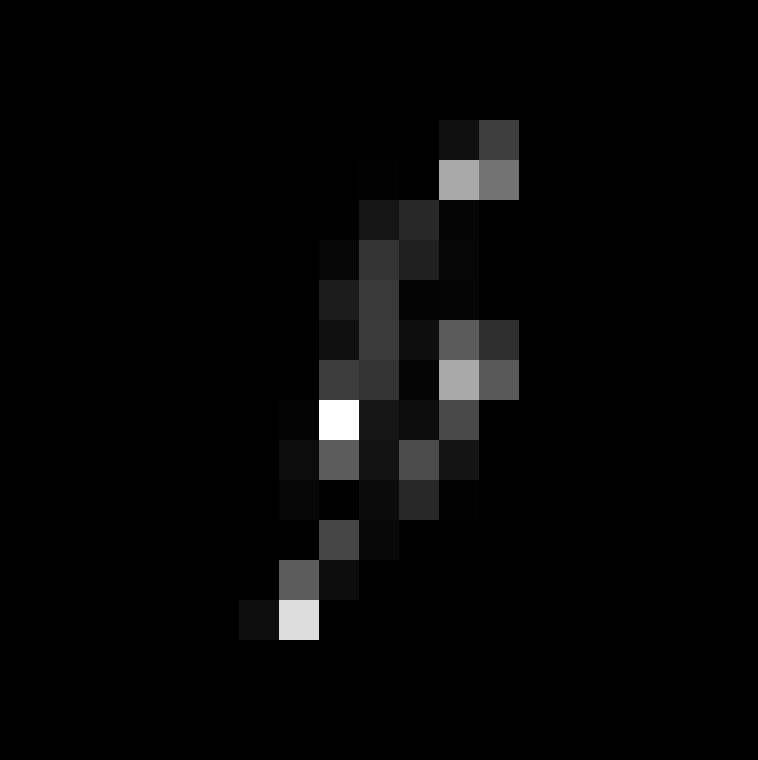} & 
      \includegraphics[width=0.100\textwidth]{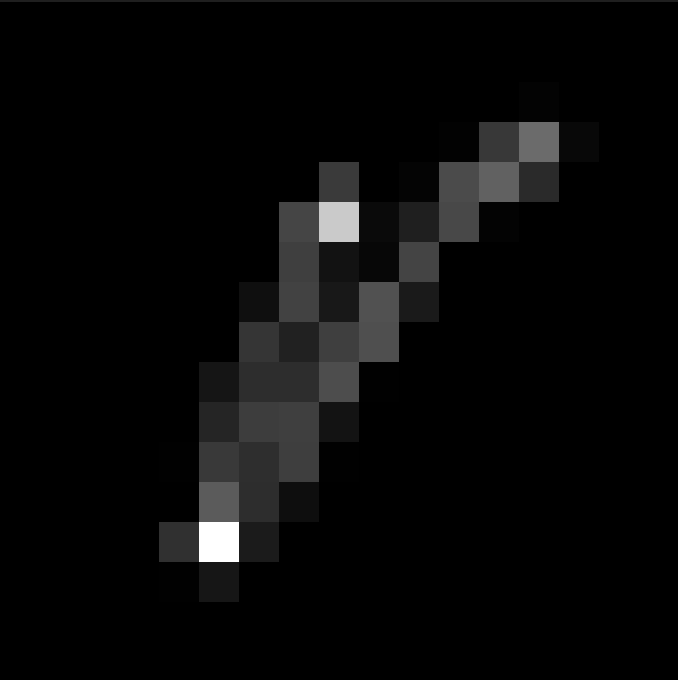} & 
       \includegraphics[width=0.100\textwidth]{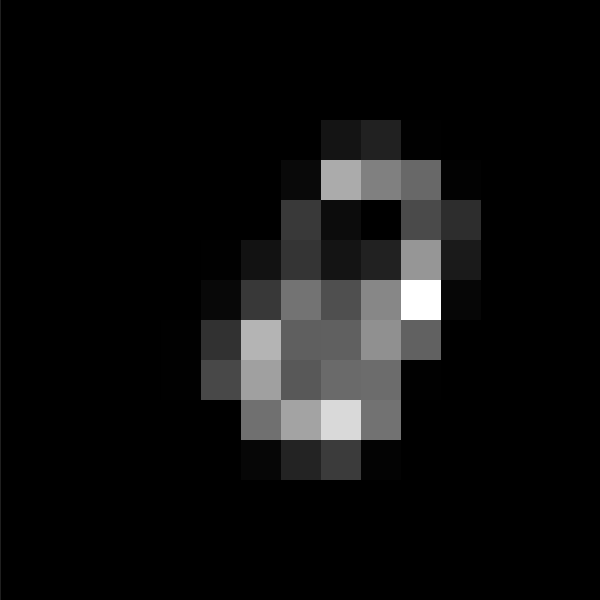} & 
        \includegraphics[width=0.100\textwidth]{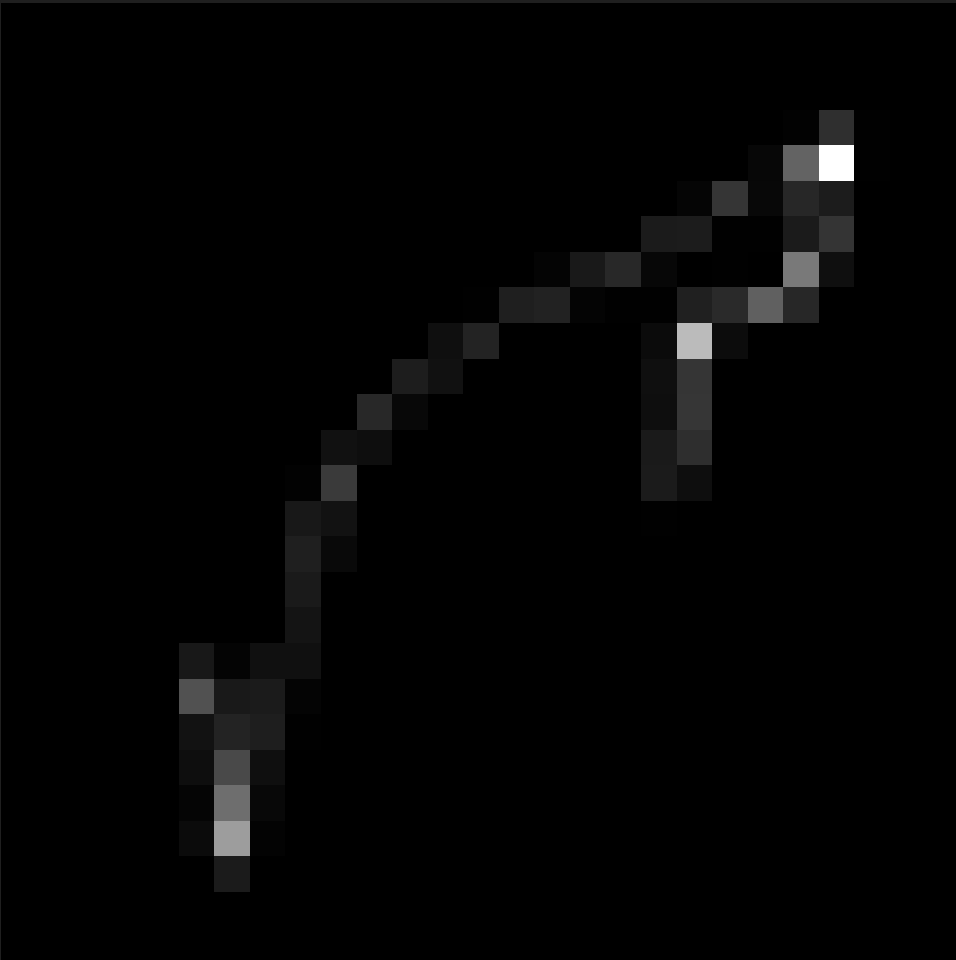}&
         \includegraphics[width=0.100\textwidth]{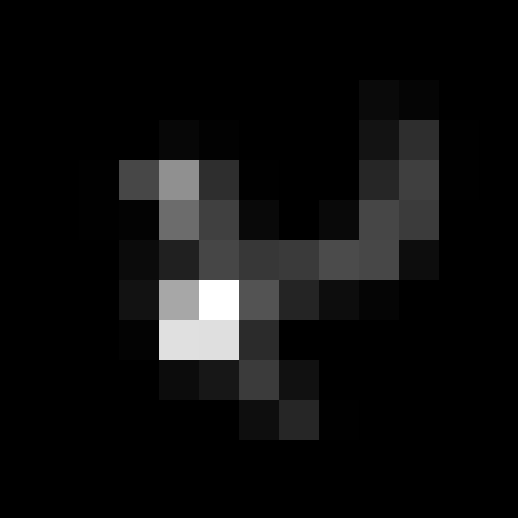} & 
      \includegraphics[width=0.100\textwidth]{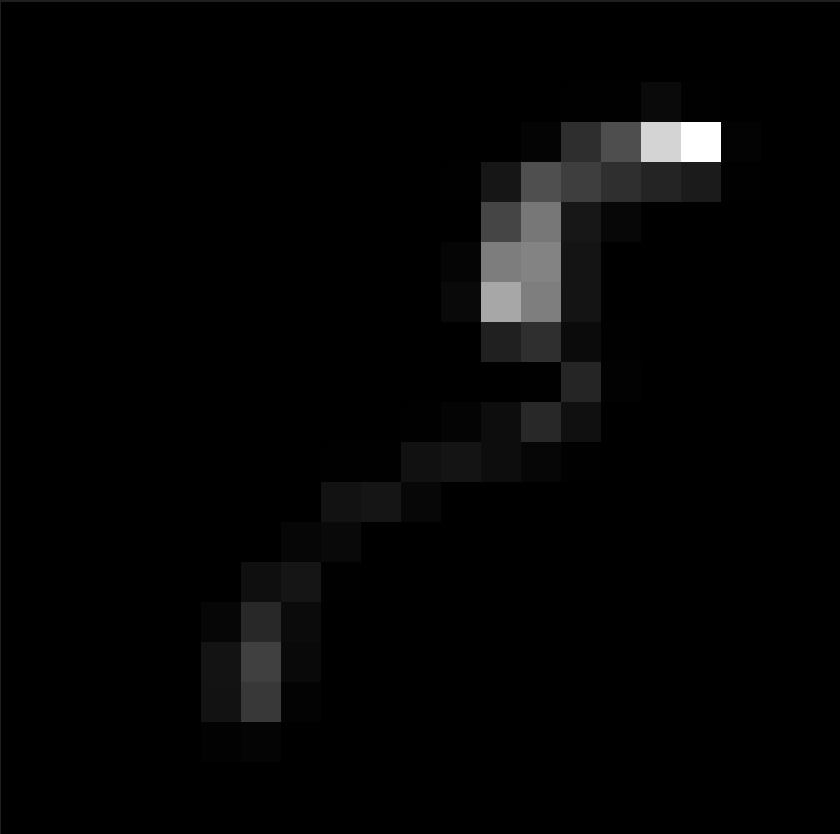} & 
       \includegraphics[width=0.100\textwidth]{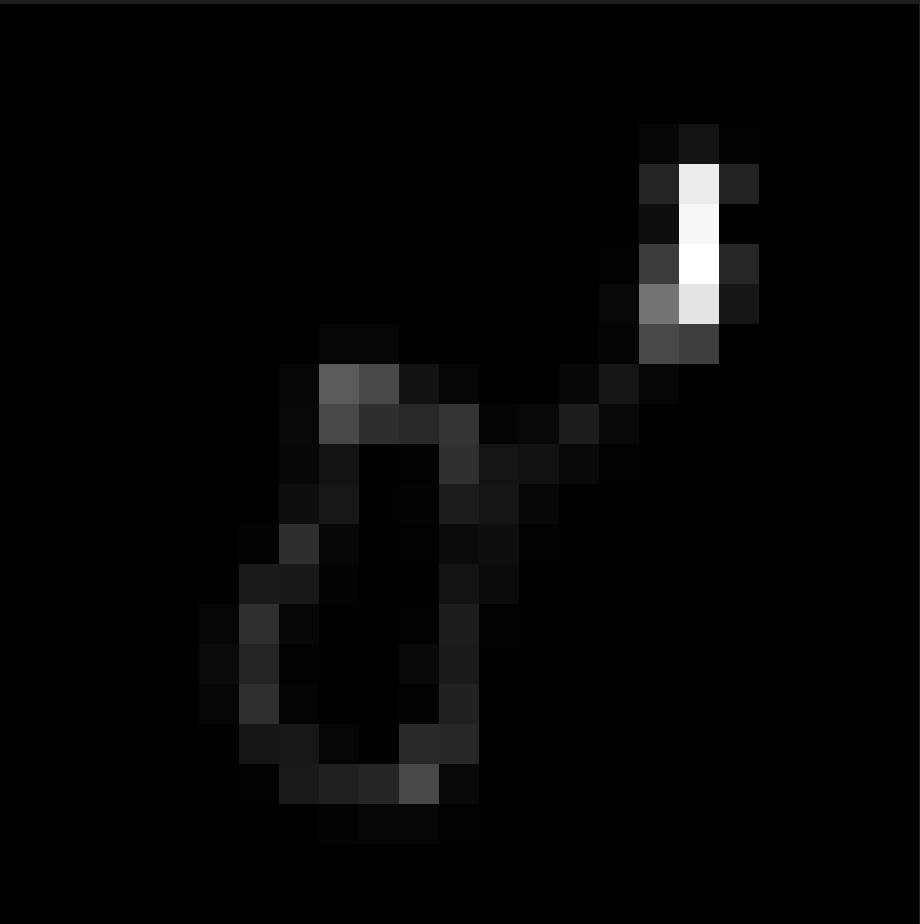} & 
        \includegraphics[width=0.100\textwidth]{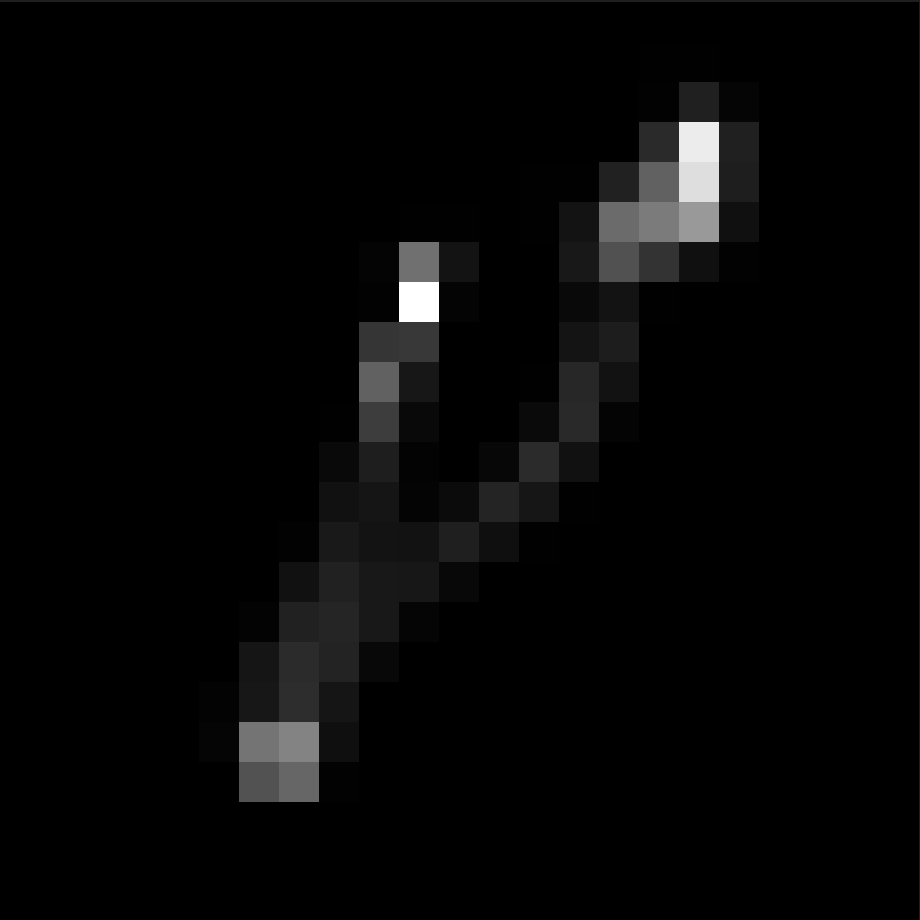}\vspace{-0.1cm}\\ 
    \end{tabular}
    }
    \caption{Kernels used for the blurring operator. These were originally shared in~\cite{Levin2009Kernels}.}
    \label{fig:mb-filters}
\end{figure}

\paragraph{Deblurring}
Deblurring is a special case of super-resolution where the scale factor is $s = 1$ and where the considered filter is no longer restricted to filtering high-frequencies (anti-aliasing) but can also be a motion blur filter for instance. This means that we consider the data-fidelity term $f(x) = \frac{1}{2}||Hx - y||^2_2$. Adapting equation \eqref{sr-Prox} to this case we also have a closed-form expression of the proximity operator of $f$:
\begin{equation}\label{deblur-Prox}
    \prox{\tau f}(z) = \mathcal{F}^*(I_M + \tau \underline{\Lambda}^* \underline{\Lambda})^{-1} \mathcal{F}(\tau H^T y + z).
\end{equation}
For deblurring we start from the observation $y$ as the initial proposal for $x_0$, we can also think of adding noise to change this initial image, but we do not consider that here.
As in~\cite{hurault2021gradient} and~\cite{ProxPnP}, we use the 8 real-world camera-shake kernels from~\cite{Levin2009Kernels} that are shown in Figure~\ref{fig:mb-filters}.
\begin{remark}
  In this paper, all convolutions are performed with periodic boundary conditions.
  This assumption is widely used in scientific imaging but is very unnatural in real-life scenarios (natural images are rarely periodic). Unfortunately when considering more realistic boundary conditions, the data-fidelity term is often not proximable.
\end{remark}

\paragraph{Inpainting}
Inpainting is another classical image restoration problem that differs from the above problems. In inpainting we apply a mask to a part of the image that hides the pixels (sets them to be all black for instance) and the goal is to infer from the visible parts a plausible full image. We can understand that this problem is highly ill-posed compared to deblurring or super-resolution that are nearly invertible (under circular boundary conditions). This gives even more importance to the regularization term. We will see that GS-PnP and Prox-PnP algorithms are suitable for solving ``random" inpainting but are not suitable for inpainting large hole.
Inpainting large holes requires to use a powerful generative model that is able to hallucinate content (see Section~\ref{sec:large-patch}). 

Another important point in inpainting is that we usually apply no noise  $y = Ax^*$.
In this case the data-fidelity term should change accordingly. We choose the convex indicator function over the reciprocal image of $y$ through $A$:
\begin{equation}
    f(x) = \iota_{A^{-1}(\{y\})}(x) = 
    \begin{cases} 
        0 &\text{if } Ax = y\\
        +\infty &\text{otherwise}.
    \end{cases}
\end{equation}
This function is non-differentiable, but it still verifies the assumption on the data-fidelity term in Theorems~\ref{Thm : 1} and \ref{Thm : 2} that guarantees  convergence for GS-PnP. Theorem \ref{Thm: DRS} requires $L < \frac{1}{2}$ to guarantee convergence of the Prox-PnP-DRS algorithm. As a consequence,  we follow~\cite{ProxPnP} and average $D_\sigma$ with $\alpha = \frac{1}{2}$. To do so, we define $D_\sigma^\alpha = \alpha D_\sigma + (1 - \alpha)\id = \id - \alpha \nabla g_\sigma$, which corresponds to taking $g_\sigma^\alpha := \alpha g_\sigma$ as the regularization function in~\eqref{phi} and using $D_\sigma^\alpha$ as the Prox-PnP denoiser.

The proximity operator of the indicator of a convex set is exactly the orthogonal projection on that set.
Here, $\prox{\tau f}(z) = \prox{A^{-1}(\{y\})}(z) = Ay - Az + z$ amounts to imposing values of $y$ outside the mask and keeping the values of $z$ inside the mask. 

Again, for inpainting we consider no noise in the observation. For random inpainting we generate random masks by using a probability $p$ of a pixel being masked, and we experimented several values $p \in \{0.1, 0.3, 0.5, 0.7, 0.9\}$.  After our experiments we notice that there is no change in algorithm behavior for different values of $p$ thus we will share plots for the fixed value $p=0.5$. We choose to initialize all masked pixels to be gray with a value of 0.5.
\paragraph{Implementation details}
The source code of our implementation heavily relies on the DeepInv open source Python library~\cite{deepinv}. The models, the parameters, the dataset, and computations such as power iteration and the proximity operators of the different data-fidelity terms are all included in this library. Our source code is based on the pseudo-codes shared in this paper.

Compared to the given pseudo-code for GS-PnP we decide here to add a stopping condition $\tau > \epsilon$ such that we do not stay too long in the backtracking process when it is activated. We choose the stopping-criterion $\epsilon = 10^{-6}$ and the maximum number of iterations to be $K = 400$. We also choose to initialize the algorithm with an image denoised by the GS-Denoiser with a large value $\sigma = 10\tau$.

For the Prox-PnP algorithms we choose $K = 1000$.

\subsubsection{Choice of $\sigma$}\label{sec:exp_sig}

\begin{figure}[!ht]
    \centering
    \resizebox{\textwidth}{!}{
    \begin{tabular}{c c c c}
 \raisebox{2em}{\makebox[0.01\textwidth]{\sidecapX{\textbf{GS-PnP\\($\lambda=0.065$)}}}}
       &   \begin{subfigure}[b]{0.3\textwidth}
           \includegraphics[width=\textwidth]{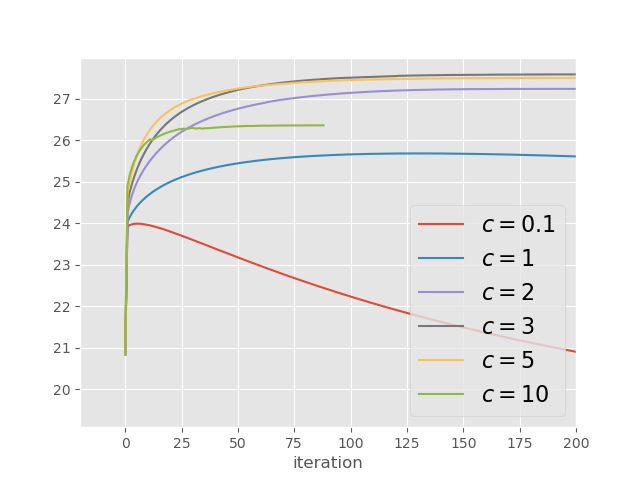}
           \caption{$\nu = 0.01$}
           \label{fig:gs-sr2:a}
       \end{subfigure} &
       \begin{subfigure}[b]{0.3\textwidth}
           \includegraphics[width=\textwidth]{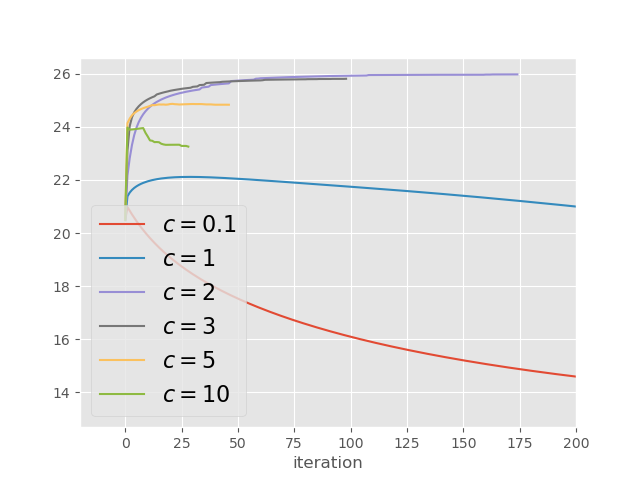}
           \caption{ $\nu = 0.03$}
           \label{fig:gs-sr2:b}
       \end{subfigure} &
       \begin{subfigure}[b]{0.3\textwidth}
           \includegraphics[width=\textwidth]{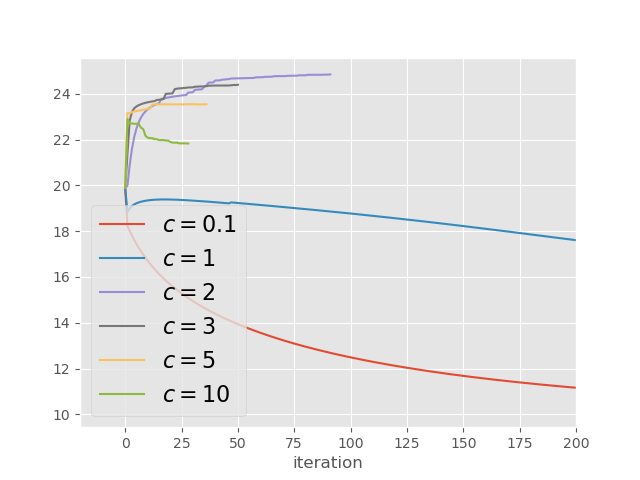}
           \caption{ $\nu = 0.05$}
           \label{fig:gs-sr2:c}
       \end{subfigure} \\
    \end{tabular}}
    \caption{\label{fig:stable-to-nu}Mean $\psnr$ evolution, for different noise levels ($\nu = 0.01$, $0.03$ and $0.05$),  over iterations of GS-PnP for $\times 2$ super-resolution with varying  parameter $c \in \{0.1, 1,2,3,5,10\}$ such that we used $\sigma = c \nu$. The mean is computed over every filter from Figure~\ref{fig:aa-filters}. For every noise level $\nu$, $\sigma =0.1\nu$ produces results with lower $\psnr$ than the initialization image. Higher $\psnr$ results are obtained for $c \in \{2, 3, 5\}$, and choosing $c$ too large results in lower $\psnr$ values (see $c = 10$). We also see that among the chosen $c \geq 2$, the GS-PnP algorithm stops in fewer iterations the higher the $\nu$ value.}
    
  \end{figure}

The parameter $\sigma$ has a strong influence on the performance of the studied algorithms as it represents the denoiser strength (how much noise it will remove). 
We choose $\sigma$ to be proportional to the observation noise level $\nu$, so we will look at different values of $\sigma = c \nu$ for varying $c$. 
As illustrated in Figure~\ref{fig:stable-to-nu}, it is the coefficient $c$ that is important, rather than the numerical value of $\sigma = c \nu$. For instance, we can see that for $\nu = 0.01$ (Figure~\ref{fig:gs-sr2:a}) the behavior of GS-PnP when $c = 5$ is different than for $\nu=0.05$ (Figure~\ref{fig:gs-sr2:c}) when $c = 1$, even though in both cases $\sigma = 0.05$. 

\begin{figure}[!ht]
    \centering
    \resizebox{\textwidth}{!}{
    \begin{tabular}{c c c}
    \textbf{GS-PnP ($\lambda = 0.065 $)} &
    \textbf{Prox-PnP-DRSdiff($\lambda = 1/0.99 $)} &
    \textbf{Prox-PnP-PGD($\lambda = 1/0.99 $)}\\
       \begin{subfigure}[b]{0.3\textwidth}
           \includegraphics[width=\textwidth]{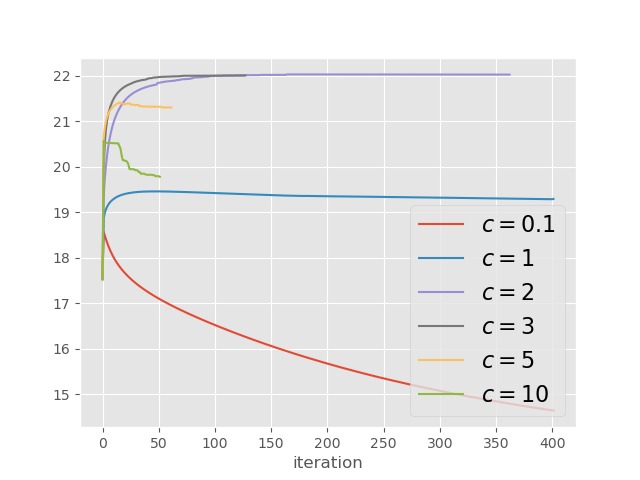}
           \caption{$\psnr$}
           \label{fig:sr4:a}
       \end{subfigure} &
       \begin{subfigure}[b]{0.3\textwidth}
           \includegraphics[width=\textwidth]{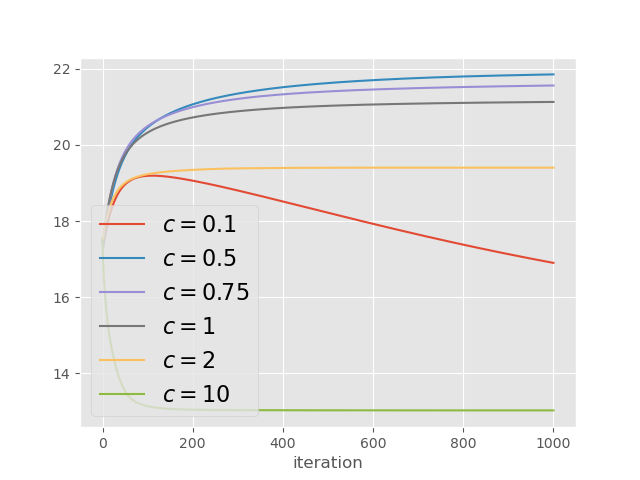}
           \caption{$\psnr$}
           \label{fig:sr4:b}
       \end{subfigure} &
       \begin{subfigure}[b]{0.3\textwidth}
           \includegraphics[width=\textwidth]{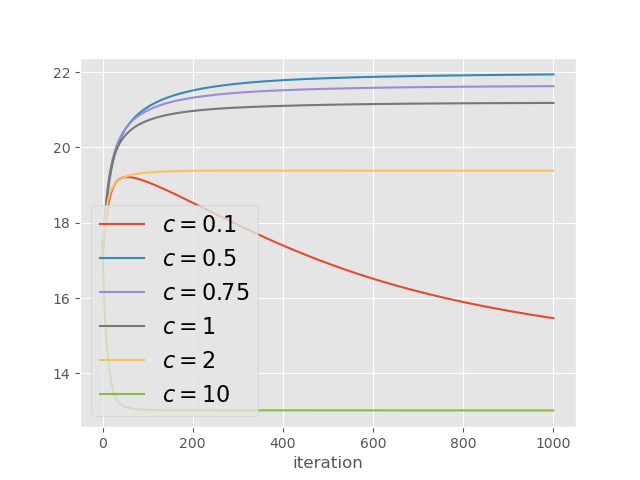}
           \caption{$\psnr$}
           \label{fig:sr4:c}
       \end{subfigure} \\
       \begin{subfigure}[b]{0.3\textwidth}
           \includegraphics[width=\textwidth]{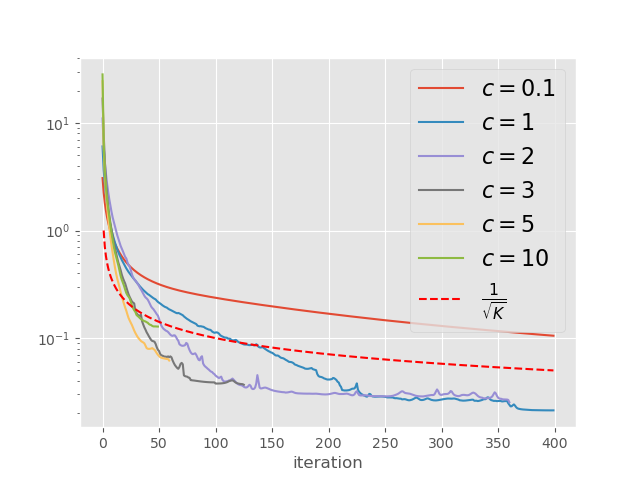}
           \caption{Residuals}
           \label{fig:sr4:d}
       \end{subfigure} &
       \begin{subfigure}[b]{0.3\textwidth}
           \includegraphics[width=\textwidth]{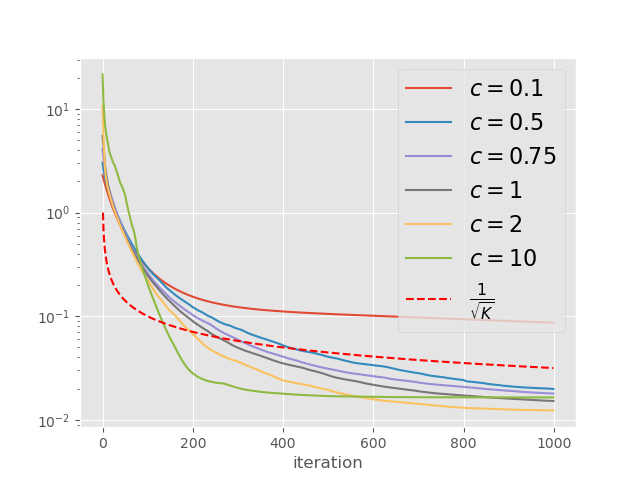}
           \caption{Residuals}
           \label{fig:sr4:e}
       \end{subfigure} &
       \begin{subfigure}[b]{0.3\textwidth}
            \includegraphics[width=\textwidth]{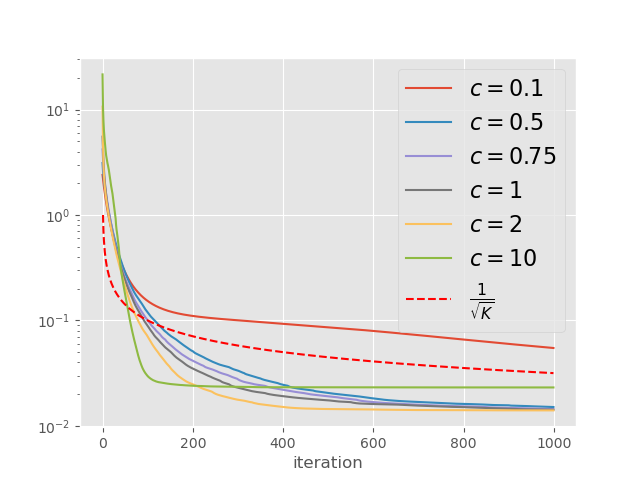}
           \caption{Residuals}
           \label{fig:sr4:f}
       \end{subfigure} \\
    \end{tabular}}
  \caption{\label{fig:sigma_sr4}Mean $\psnr$ and residual  evolution over iterations of each algorithm for $\times 4$ super-resolution with varying  parameter $c$ and $\nu = 0.05$ such that $\sigma = c \nu$. The mean is computed over every anti-aliasing filter from Figure~\ref{fig:aa-filters}. We see in the $\psnr$ plots (\ref{fig:sr4:a}, \ref{fig:sr4:b}, \ref{fig:sr4:c}) that there is a bounded interval of coefficients $c$ to choose to obtain high $\psnr$ results. We also notice in (\ref{fig:sr4:d}, \ref{fig:sr4:e}, \ref{fig:sr4:f}) that the algorithms follow the announced convergence rate of the residuals of $O(\frac{1}{\sqrt{K}})$. 
  }
    
\end{figure}

In the  $\psnr$ plots of Figure~\ref{fig:sigma_sr4}, we show that for each algorithm there is a bounded interval for coefficients $c$ that result in higher $\psnr$ solutions.
Choosing $c$ too small or too large gives results with lower $\psnr$ than the initially proposed image.

For every chosen $\sigma$ the algorithm converges to a solution.
It is shown on the residual plots of Figure~\ref{fig:sigma_sr4} that the residuals converge correctly to $0$ and with the announced rate of $O(\frac{1}{\sqrt{K}})$ with $K$ the number of iterations. 
We also notice in Figure~\ref{fig:sr4:a} that the higher $c$, the faster the stopping criterion is met for the GS-PnP algorithm.

\begin{figure}[!ht]
\centering
    \begin{tabular}{c c}
    \textbf{observation}  & \textbf{groundtruth}\\
    \begin{overpic}[width=0.33\textwidth]{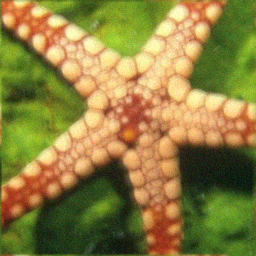} 
        \put(0, 69){\includegraphics[width=0.1\textwidth]{images/filters/blur-filter_4.png}} 
    \end{overpic} & 
        \includegraphics[width=0.33\textwidth]{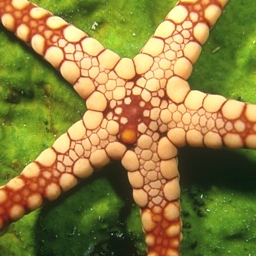}
        \\
    \end{tabular}
    \resizebox{\textwidth}{!}{
    \begin{tabular}{@{}c c c c}
        \raisebox{4em}{\makebox[0.03\textwidth]{\sidecapX{\textbf{Prox-PnP-DRSdiff ($\lambda=0.99$)}}}}
        &
        \begin{subfigure}[b]{0.33\textwidth}
           \includegraphics[width=\textwidth]{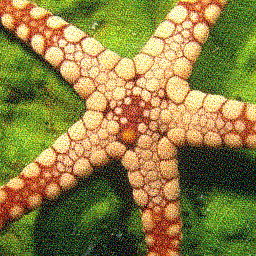}
           \caption{\textbf{$\sigma = 0.1 \nu$}\\
           $\psnr=17.32$}
           \label{im:deb:a}
       \end{subfigure}
         &
         \begin{subfigure}[b]{0.33\textwidth}
           \includegraphics[width=\textwidth]{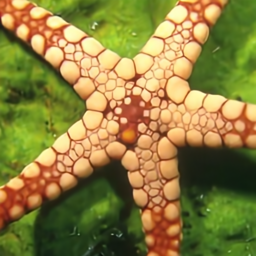}
           \caption{\textbf{$\sigma = 0.5 \nu$}\\
           $\psnr=31.16$}
           \label{im:deb:b}
       \end{subfigure}
       &
       \begin{subfigure}[b]{0.33\textwidth}
           \includegraphics[width=\textwidth]{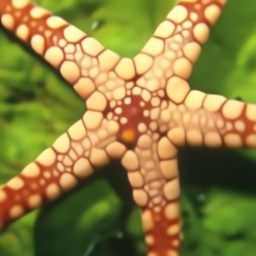}
           \caption{\textbf{$\sigma = 2 \nu$}\\
           $\psnr=27.79$}
           \label{im:deb:c}
       \end{subfigure}\\
    \end{tabular}}
    \caption{\label{fig:images_deb}Deblurring results after 1000 iterations of the Prox-PnP-DRSdiff algorithm with varying $\sigma$ parameter for the Prox-denoiser with $\nu = 0.03$. We can see the effect of the choice of $\sigma$, for a value that is too small the resulting image is very sharp and artifacts appear (a) whereas a $\sigma$ that is too big causes unwanted smoothing as in (c) (see background and starfish legs).}
    
\end{figure}

Figure~\ref{fig:images_deb} illustrates the effect of $\sigma$ on the deblurring result.
For GS-PnP we choose the same arbitrary $\lambda$ as that of the original paper~\cite{hurault2021gradient}: $\lambda = 0.065$. For Prox-PnP-PGD and Prox-PnP-DRSdiff as seen in Theorems~\ref{Thm: PGD} and~\ref{Thm: DRSdiff} we must verify $\lambda > L_f$, where $L_f$ is the Lipschitz coefficient of the gradient of the data-fidelity term $f(x) = \frac{1}{2}||Ax - y||^2$. With this choice of $f$ we have $L_f = |||A^T A|||$.
In both super-resolution and deblurring cases,
one can show that $L_f = |||H^T H||| = 1$.
In order to regularize sufficiently while meeting the above condition, we choose $\lambda = 1/0.99$.

We also note that in our experiments, the Prox-PnP-PGD and Prox-PnP-DRSdiff will converge towards the minimizers of the same objective function $F^{\lambda,\sigma}$.

\subsubsection{Choice of $\lambda$}
\begin{figure}[!ht]
\centering
    \begin{tabular}{c c}
    \textbf{observation}  & \textbf{groundtruth}\\
    \includegraphics[width=0.28\textwidth]{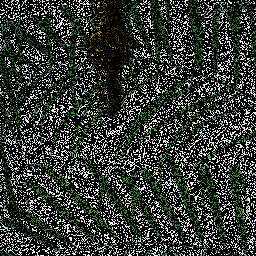}
    & 
        \includegraphics[width=0.28\textwidth]{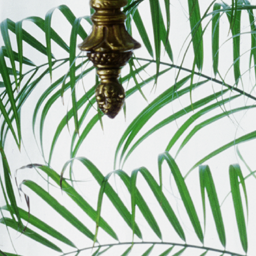}
        \\
    \end{tabular}

    \begin{tabular}{@{}c c c c}
        \raisebox{3em}{\makebox[0.03\textwidth]{\sidecapX{\textbf{GS-PnP ($\sigma=30/255$)}}}}
        &
        \begin{subfigure}[b]{0.28\textwidth}
           \includegraphics[width=\textwidth]{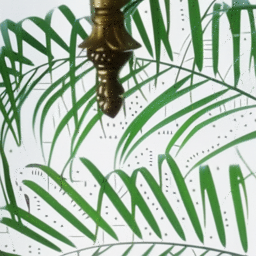}
           \caption{\small\textbf{$\lambda = 0.1$}\\
           $\psnr=25.43$}
           \label{fig:inp:a}
       \end{subfigure}
         &
        \begin{subfigure}[b]{0.28\textwidth}
           \includegraphics[width=\textwidth]{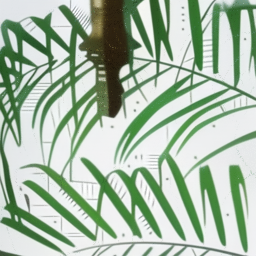}
           \caption{\textbf{$\lambda = 2$}\\
           $\psnr=23.92$}
           \label{fig:inp:b}
       \end{subfigure}
       &
        \begin{subfigure}[b]{0.28\textwidth}
           \includegraphics[width=\textwidth]{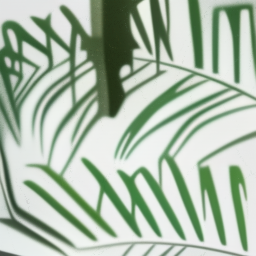}
           \caption{\textbf{$\lambda = 10$}\\
           $\psnr=19.92$}
           \label{fig:inp:c}
       \end{subfigure}
       \\
    \end{tabular}
    \caption{\label{fig:images_inp}Random masked inpainting results after 200 iterations of the GS-PnP algorithm with varying regularization parameter $\lambda$ with $p = 0.7$. We can see the effect of the choice of $\lambda$: low regularization leads to sharper results with more details, but may keep some masked pixels~(\ref{fig:inp:a}) whereas higher regularization reduces details but inpaints all masked pixels~(\ref{fig:inp:c}). It seems that it is hard to find a parameter that gives results without visual artifacts for this large $p$. As we can see for $\lambda = 2$ the result is not detailed enough but is already polluted by masked pixels~(\ref{fig:inp:b}).}
\end{figure}

\begin{figure}[!ht]
    \centering
    \begin{tabular}{c c c}
    \textbf{GS-PnP ($\sigma = 1.8\nu$)} &
    \textbf{Prox-PnP-DRSdiff($\sigma = 0.5\nu$)} &
    \textbf{Prox-PnP-PGD($\sigma = 0.5\nu$)} \\
       \begin{subfigure}[b]{0.305\textwidth}
           \includegraphics[width=\textwidth]{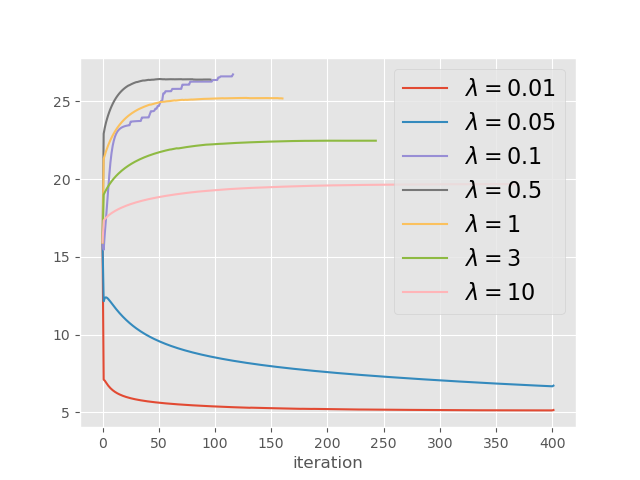}
           \caption{$\psnr$}
           \label{fig:deb:a}
       \end{subfigure} &
       \begin{subfigure}[b]{0.305\textwidth}
           \includegraphics[width=\textwidth]{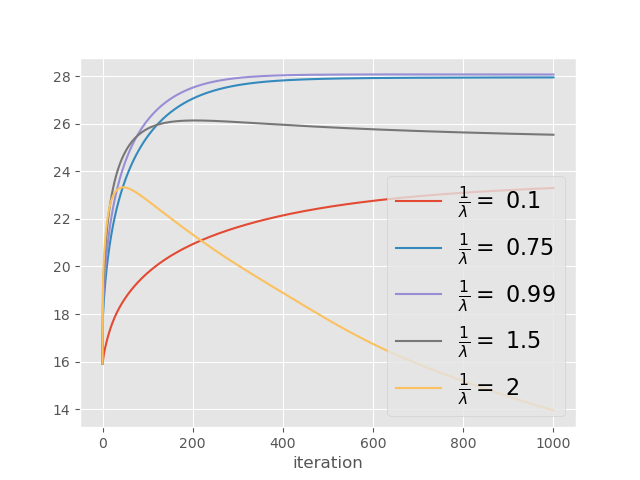}
           \caption{$\psnr$}
           \label{fig:deb:b}
       \end{subfigure} &
       \begin{subfigure}[b]{0.305\textwidth}
           \includegraphics[width=\textwidth]{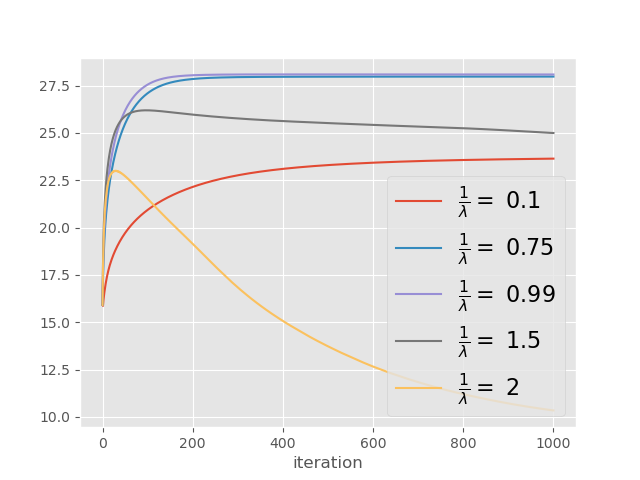}
           \caption{$\psnr$}
           \label{fig:deb:c}
       \end{subfigure} \\
       \begin{subfigure}[b]{0.305\textwidth}
           \includegraphics[width=\textwidth]{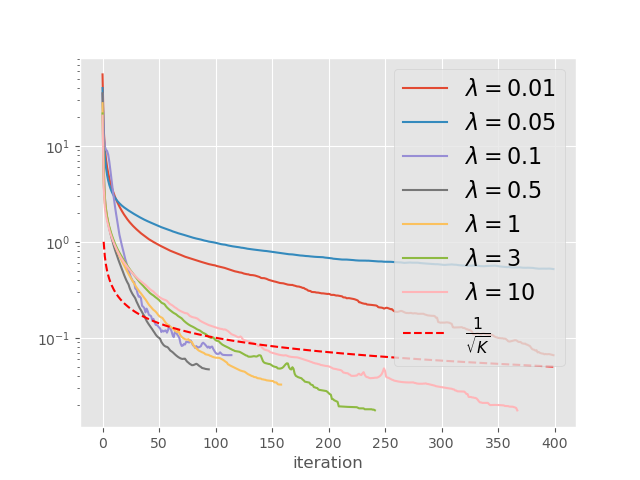}
           \caption{Residuals}
           \label{fig:deb:d}
       \end{subfigure} &
       \begin{subfigure}[b]{0.305\textwidth}
           \includegraphics[width=\textwidth]{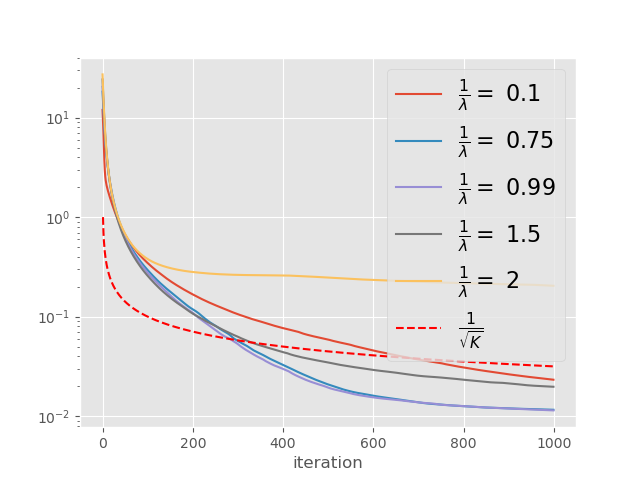}
           \caption{Residuals}
           \label{fig:deb:e}
       \end{subfigure} &
       \begin{subfigure}[b]{0.305\textwidth}
           \includegraphics[width=\textwidth]{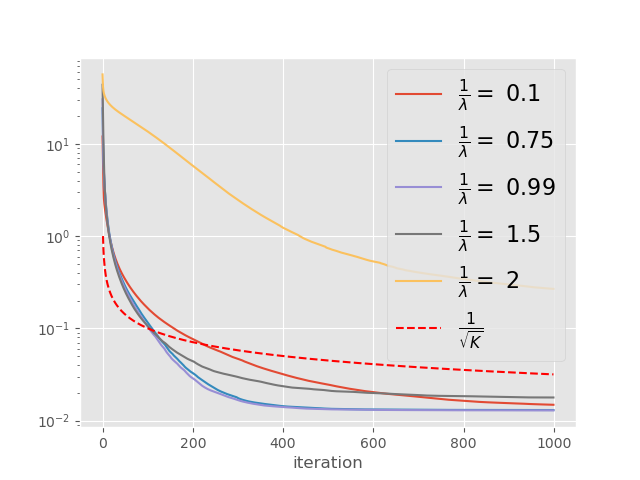}
           \caption{Residuals}
           \label{fig:deb:f}
       \end{subfigure} \\
    \end{tabular}
  \caption{\label{fig:exp_lam}Mean $\psnr$ and residuals  evolution over iterations of each algorithm for deblurring with varying $\lambda$ parameter and $\nu = 0.05$. The mean is computed over every motion blur filter from Figure~\ref{fig:mb-filters}.
    We see that $\lambda$ must be carefully tuned in order to optimize visual performance.
    These results again confirm the convergence speed of $O(\frac{1}{\sqrt{K}})$ (see~\ref{fig:deb:d}, \ref{fig:deb:e}, \ref{fig:deb:f}).}
    
\end{figure}

The parameter $\lambda$  sets the balance between the data-fidelity and regularization terms. In our formulation the amount of regularization is inversely proportional to $\lambda$.
In the PnP setting this allows to adjust the constraint of the output image to lie in the denoiser training domain of noiseless images. This interpretation is not very far from what we interpret from $\sigma$ in Section~\ref{sec:exp_sig}, since denoising causes also some form of smoothing, we expect $\lambda$ to also smooth out images the more we regularize.
Therefore, increasing $\lambda$ has a similar effect than increasing $\sigma$, as can be observed in Figure~\ref{fig:images_inp}.
Convergence monitoring and $\psnr$ values can be seen in Figure~\ref{fig:exp_lam}. 
\subsubsection{Choice of $\tau_0$}
\begin{figure}[!ht]
    \centering

    \begin{tabular}{c c c}
           \raisebox{2em}{\makebox[0.01\textwidth]{\sidecapX{\textbf{GS-PnP\\($\lambda=0.065$)}}}}
       &   \begin{subfigure}[b]{0.4\textwidth}
           \includegraphics[width=\textwidth]{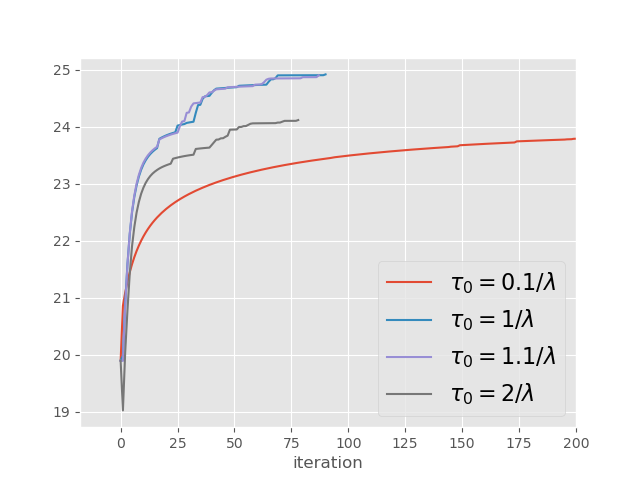}
           \caption{$\psnr$}
           \label{fig:tau-sr2:a}
       \end{subfigure} &
       \begin{subfigure}[b]{0.4\textwidth}
           \includegraphics[width=\textwidth]{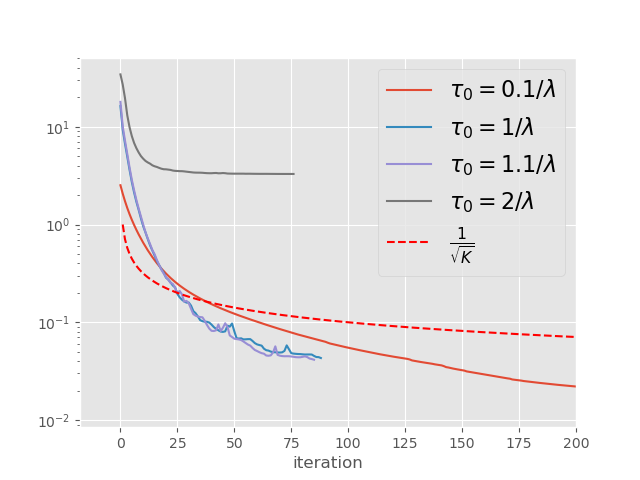}
           \caption{Residuals}
           \label{fig:tau-sr2:b}
       \end{subfigure} \\
    \end{tabular}
    \caption{Mean $\psnr$ and residuals  evolution over iterations of GS-PnP for $\times 2$ super-resolution with varying initial step size $\tau_0$. The mean is computed over every filter from Figure~\ref{fig:aa-filters}. As expected the larger the initial step size, the faster the algorithm stopping criterion is reached~\ref{fig:tau-sr2:a}. That said we must choose one $\tau_0$ that is large for fast convergence, but not too large for performance. The graph~\ref{fig:tau-sr2:b} also shows the impact of choosing a large step size, the residual values are higher than for smaller step sizes since the iterates are further appart. Even if we have a  theoretical guarantee that the residuals converge to 0, larger initial step sizes slow down this convergence speed.}
    \label{fig:tau0_exp}
\end{figure}

\begin{figure}[!ht]
\centering

    \begin{tabular}{c c}
  \textbf{observation}  & \textbf{groundtruth}\\
    \begin{overpic}[width=0.14\textwidth]{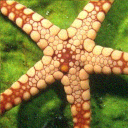} 
        \put(0, 65){\includegraphics[width=0.05\textwidth]{images/filters/filter_0.png}} 
    \end{overpic} & 
        \includegraphics[width=0.28\textwidth]{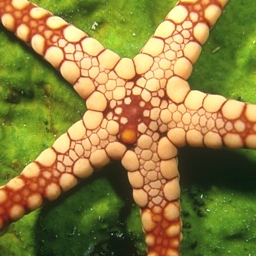}
    \end{tabular}
    \begin{tabular}{@{}c c c c}
        \raisebox{4.6em}{\makebox[0.03\textwidth]{\sidecapX{\textbf{GS-PnP ($\lambda=0.065, \sigma = 2\nu$)}}}}
        &
        \begin{subfigure}[b]{0.28\textwidth}
           \includegraphics[width=\textwidth]{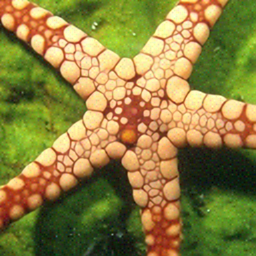}
           \caption{\textbf{$\tau_0 = 0.1 / \lambda$,  $\psnr=28.88$}\\
          num.iter.= 143\vspace*{-.2cm}}
           \label{im:sr2:a}
       \end{subfigure}
         &
         \begin{subfigure}[b]{0.28\textwidth}
           \includegraphics[width=\textwidth]{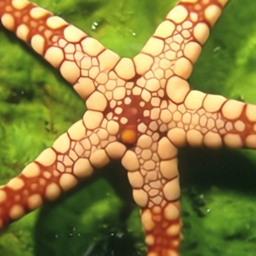}
           \caption{\textbf{$\tau_0 = 1 / \lambda$, $\psnr=31.03$}\\
                 num.iter.= 35\vspace*{-.2cm}}
           \label{im:sr2:b}
       \end{subfigure}
       &
       \begin{subfigure}[b]{0.28\textwidth}
           \includegraphics[width=\textwidth]{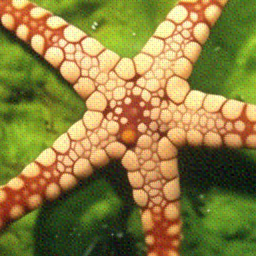}
           \caption{\textbf{$\tau_0 = 2 / \lambda$, $\psnr=25.60$}\\
           num.iter.= 2\vspace*{-.2cm}}
           \label{im:sr2:c}
       \end{subfigure}\\
    \end{tabular}
    \caption{\label{fig:images_sr2}$\times 2$ Super-resolution results from the GS-PnP algorithm with varying initial step size $\tau_0$ with $\nu = 0.03$. We can see the effect of the choice of $\tau_0$, for a value that is small the resulting image~(\ref{im:sr2:a}) is close to best performance~(\ref{im:sr2:b}) but the stopping criterion is reached in a large number of steps, this is expected when using a small step size. Choosing an initial step size that is too large~(\ref{im:sr2:c}) causes very early stopping with worsening results as $\tau_0$ increases.}
    
\end{figure}

These last experiments only concern the GS-PnP algorithm, which converges for $\tau < \frac{1}{\lambda L}$ where $L$ is the Lipschitz constant of $\nabla g_\sigma$.
We do recall that $\tau$ as a step size determines the convergence speed of an algorithm. Taking small steps will lead to reaching the solution in a higher amount of steps whereas choosing a large step size should allow to reach it in less steps.
As explained in Section~\ref{backtracking} the backtracking procedure makes sure that we have sufficient decrease at each iteration.
As discussed in~\cite{hurault2021gradient} in practice for the majority of images 
we may choose $\tau_0 \approx \lambda$, but here we will try other values of $\tau_0$.
Figures~\ref{fig:tau0_exp} and~\ref{fig:images_sr2} show the impact of the initial step size $\tau_0$ on the resulting $\psnr$ and the convergence speed of the GS-PnP algorithm. 
The initial value of $\tau_0$ has a major impact on the beginning of the algorithm, which may not be corrected afterwards.
Unfortunately we can sometimes ``step over" the solution. 

\subsection{Difficulty with large band inpainting}\label{sec:large-patch}
\begin{figure}[!ht]
\centering
    \resizebox{\textwidth}{!}{
    \begin{tabular}{c c c c c}
    \raisebox{2.em}{\makebox[0.025\textwidth]{\sidecapX{\hspace{-0.3cm}\textbf{Prox-PnP-DRS\newline \hspace*{-0.7cm}($\lambda~=~1000, \sigma~=~75 /255$)}}}} &
    \begin{subfigure}[b]{0.24\textwidth}
           \includegraphics[width=\textwidth]{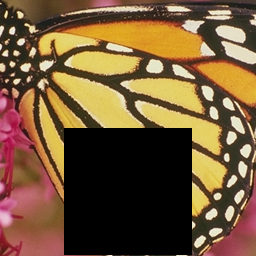}
           \caption{Observation\vspace*{-.2cm}}
           \label{fig:band:a}
       \end{subfigure} &
       \begin{subfigure}[b]{0.24\textwidth}
           \includegraphics[width=\textwidth]{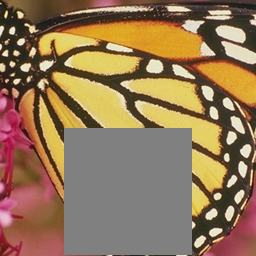}
           \caption{ Initial ($\psnr=17.32$)\vspace*{-.2cm}}
           \label{fig:band:b}
       \end{subfigure} 
        &
         \begin{subfigure}[b]{0.24\textwidth}
           \includegraphics[width=\textwidth]{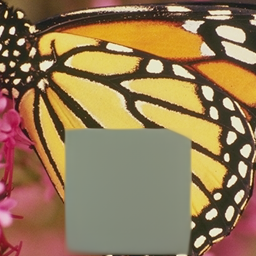}
           \caption{Result ($\psnr=17.39$)\vspace*{-.2cm}}
           \label{fig:band:c}
        \end{subfigure} & \begin{subfigure}[b]{0.24\textwidth}
           \includegraphics[width=\textwidth]{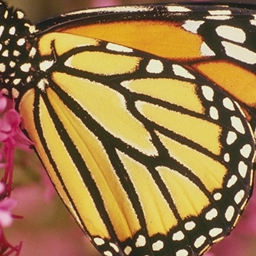}
           \caption{Groundtruth\vspace*{-.2cm}}
           \label{fig:band:d}
       \end{subfigure}
       \\
    \end{tabular}}
    \caption{\label{fig:large_band}Large area inpainting example. PnP methods cannot generate structured information and inpaints only smooth colors at the boundary of the mask.}
   
\end{figure}

Large area inpainting is a difficult task for the studied plug-and-play algorithms, as can be observed on Figure~\ref{fig:large_band}.
After 200 iterations, the only effect of the algorithm is to slightly change the regions color to get closer to the outer area color (yellow).
The corners are slightly filled, with relatively smooth edges.
In fact, these plug-and-play algorithms cannot hallucinate content (border or texture). 
Also it is hard for algorithms to determine whether a uniformly colored area is natural or not (shadows, white walls) and thus if it should be inpainted or not. 

\vspace*{1cm}

\section{Discussion and Conclusion}\label{sec:conclusion}
In this work we have presented the Gradient-Step Plug-and-Play framework from~\cite{hurault2021gradient} and~\cite{ProxPnP}.
We have shown the reproducibility of these algorithms, and studied the influence of the parameters of interest.
We noticed that the parameters have an interpretable usage, that these behaviours remain the same at several noise levels and for different problems.
However we have illustrated that such Plug-and-Play methods cannot be used for severely ill-posed problems (e.g. inpainting large holes). On the one hand, they cannot hallucinate content in order to cope with so ill-posed problems; on the other hand they offer relatively stable behavior on more common inverse problems.

\section*{Acknowledgment}
This work benefited from the support of the projects  HoliBrain of the French National Research Agency (ANR-23-CE45-0020-01).
\section*{Image Credits}

{\small
\includegraphics[height=2em]{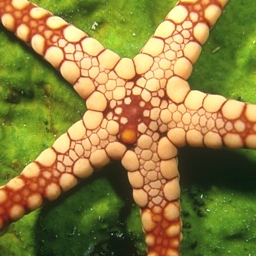}
\includegraphics[height=2em]{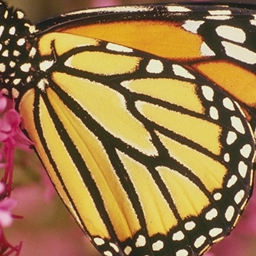}
\includegraphics[height=2em]{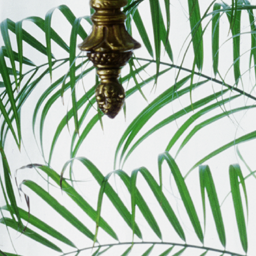}
- Set3C dataset. \\
}
{\small\includegraphics[height=3em]{images/DRUNET.png} - Fig.1 from \cite{DRUnet}, permission to use given by Kai Zhang.}


\bibliographystyle{siam}

\begin{thebibliography}{10}

\bibitem{DIV2K}
{\sc E.~Agustsson and R.~Timofte}, {\em Ntire 2017 challenge on single image
  super-resolution: Dataset and study}, in 2017 IEEE Conference on Computer
  Vision and Pattern Recognition Workshops (CVPRW), 2017, pp.~1122--1131.

\bibitem{KL}
{\sc H.~Attouch, J.~Bolte, P.~Redont, and A.~Soubeyran}, {\em Proximal
  alternating minimization and projection methods for nonconvex problems. an
  approach based on the kurdyka-lojasiewicz inequality}, 2013.

\bibitem{Beck}
{\sc A.~Beck}, {\em First-Order Methods in Optimization}, Society for
  Industrial and Applied Mathematics, Philadelphia, PA, 2017.

\bibitem{deepinv}
{\em {DeepInverse: A deep learning framework for inverse problems in imaging}},
  2023.
\newblock https://deepinv.github.io/deepinv/.

\bibitem{Tweedie}
{\sc B.~Efron}, {\em Tweedie's formula and selection bias}, Journal of the
  American Statistical Association, 106 (2011), pp.~1602--1614.

\bibitem{MAP}
{\sc R.~Gribonval}, {\em Should penalized least squares regression be
  interpreted as maximum a posteriori estimation?}, IEEE Transactions on Signal
  Processing, 59 (2011), pp.~2405--2410.

\bibitem{resnet}
{\sc K.~He, X.~Zhang, S.~Ren, and J.~Sun}, {\em Deep residual learning for
  image recognition}, 2015.

\bibitem{hurault2023thesis}
{\sc S.~Hurault}, {\em Convergent plug-and-play methods for image inverse
  problems with explicit and nonconvex deep regularization}, PhD thesis,
  Universit{\'e} de Bordeaux, 2023.

\bibitem{hurault2021gradient}
{\sc S.~Hurault, A.~Leclaire, and N.~Papadakis}, {\em Gradient step denoiser
  for convergent plug-and-play}, in International Conference on Learning
  Representations, 2021.

\bibitem{ProxPnP}
\leavevmode\vrule height 2pt depth -1.6pt width 23pt, {\em Proximal denoiser
  for convergent plug-and-play optimization with nonconvex regularization}, in
  International Conference on Machine Learning, 2022.

\bibitem{Levin2009Kernels}
{\sc A.~Levin, Y.~Weiss, F.~Durand, and W.~T. Freeman}, {\em Understanding and
  evaluating blind deconvolution algorithms}, 2009 IEEE Conference on Computer
  Vision and Pattern Recognition,  (2009), pp.~1964--1971.

\bibitem{FLICK2K}
{\sc B.~Lim, S.~Son, H.~Kim, S.~Nah, and K.~M. Lee}, {\em Enhanced deep
  residual networks for single image super-resolution}, 2017.

\bibitem{PnP-eig}
{\sc J.~Liu, M.~S. Asif, B.~Wohlberg, and U.~S. Kamilov}, {\em Recovery
  analysis for plug-and-play priors using the restricted eigenvalue condition},
  2021.

\bibitem{WED}
{\sc K.~Ma, Z.~Duanmu, Q.~Wu, Z.~Wang, H.~Yong, H.~Li, and L.~Zhang}, {\em
  Waterloo exploration database: New challenges for image quality assessment
  models}, IEEE Transactions on Image Processing, 26 (2017), pp.~1004--1016.

\bibitem{wavelet}
{\sc S.~Mallat}, {\em A Wavelet Tour of Signal Processing, Third Edition: The
  Sparse Way}, Academic Press, Inc., USA, 3rd~ed., 2008.

\bibitem{CBSD}
{\sc D.~Martin, C.~Fowlkes, D.~Tal, and J.~Malik}, {\em A database of human
  segmented natural images and its application to evaluating segmentation
  algorithms and measuring ecological statistics}, in IEEE International
  Conference on Computer Vision. ICCV 2001, vol.~2, 2001, pp.~416--423 vol.2.

\bibitem{spec-norm}
{\sc J.-C. Pesquet, A.~Repetti, M.~Terris, and Y.~Wiaux}, {\em Learning
  maximally monotone operators for image recovery}, 2021.

\bibitem{RED-Clar}
{\sc E.~T. Reehorst and P.~Schniter}, {\em Regularization by denoising:
  Clarifications and new interpretations}, IEEE Transactions on Computational
  Imaging, 5 (2019), pp.~52--67.

\bibitem{renaud2025stability}
{\sc M.~Renaud, V.~De~Bortoli, A.~Leclaire, and N.~Papadakis}, {\em From
  stability of {L}angevin diffusion to convergence of proximal {MCMC} for
  non-log-concave sampling}, arXiv preprint arXiv:2505.14177,  (2025).

\bibitem{RED}
{\sc Y.~Romano, M.~Elad, and P.~Milanfar}, {\em The little engine that could:
  Regularization by denoising (red)}, 2017.

\bibitem{Unet}
{\sc O.~Ronneberger, P.~Fischer, and T.~Brox}, {\em U-net: Convolutional
  networks for biomedical image segmentation}, 2015.

\bibitem{TV}
{\sc L.~I. Rudin, S.~Osher, and E.~Fatemi}, {\em Nonlinear total variation
  based noise removal algorithms}, Physica D: Nonlinear Phenomena, 60 (1992),
  pp.~259--268.

\bibitem{PnP-PTD}
{\sc E.~Ryu, J.~Liu, S.~Wang, X.~Chen, Z.~Wang, and W.~Yin}, {\em Plug-and-play
  methods provably converge with properly trained denoisers}, in International
  Conference on Machine Learning, vol.~97, PMLR, 2019, pp.~5546--5557.

\bibitem{Scalable-PnP}
{\sc Y.~Sun, Z.~Wu, X.~Xu, B.~Wohlberg, and U.~S. Kamilov}, {\em Scalable
  plug-and-play admm with convergence guarantees}, IEEE Transactions on
  Computational Imaging, 7 (2021), pp.~849--863.

\bibitem{Tikhonov:1963}
{\sc A.~N. Tikhonov}, {\em Solution of incorrectly formulated problems and the
  regularization method}, Soviet Math. Dokl., 4 (1963), pp.~1035--1038.

\bibitem{PnP}
{\sc S.~V. Venkatakrishnan, C.~A. Bouman, and B.~Wohlberg}, {\em Plug-and-play
  priors for model based reconstruction}, in 2013 IEEE Global Conference on
  Signal and Information Processing, 2013, pp.~945--948.

\bibitem{DRUnet}
{\sc K.~Zhang, Y.~Li, W.~Zuo, L.~Zhang, L.~V. Gool, and R.~Timofte}, {\em
  Plug-and-play image restoration with deep denoiser prior}, 2021.

\bibitem{prox-sr}
{\sc N.~Zhao, Q.~Wei, A.~Basarab, N.~Dobigeon, D.~Kouam{\'e}, and J.-Y.
  Tourneret}, {\em Fast single image super-resolution using a new analytical
  solution for $\ell_2-\ell_2$ problems}, IEEE Transactions on Image
  Processing, 25 (2016), pp.~3683--3697.

\end{thebibliography}

\end{document}